\documentclass[10pt,a4paper,twocolumn]{article}

\usepackage{graphicx} %
\usepackage{url}
\usepackage{tabularx}
\usepackage{hyperref}
\usepackage{algorithm}
\usepackage{algpseudocode}
\usepackage{placeins}
\usepackage{booktabs}
\usepackage{stfloats}

\usepackage{todonotes}

\usepackage{subcaption}

\usepackage{dcolumn}           %
\newcolumntype{.}   {D{.}{.}{-1}} %
\newcolumntype{d}[1]{D{.}{.}{#1}} %
\newcolumntype{e}   {D{E}{E}{-1}} %
\newcolumntype{E}[1]{D{E}{E}{#1}} %
\newcolumntype{C}{>{\centering\arraybackslash}X}

\usepackage{amsmath}
\usepackage{mathptmx}

\usepackage{amssymb}
\usepackage{amsthm}

\usepackage{float}

\usepackage{setspace}

\usepackage{sectsty}

\newcommand{\myFontSize}{\fontsize{10}{12}\selectfont}

\sectionfont{\myFontSize}       %
\subsectionfont{\rm\myFontSize\itshape} %

\usepackage{titlesec}
\titlespacing*{\section}{0pt}{10pt}{0pt}
\titlespacing*{\subsection}{0pt}{10pt}{0pt}

\usepackage{secdot}
\sectiondot{subsection}
\sectionpunct{section}{. }    %
\sectionpunct{subsection}{. } %

\usepackage{caption}
\usepackage{xcolor} %
\usepackage{fancyhdr} %
\usepackage{booktabs}
\usepackage{lastpage}
\hypersetup{
    colorlinks,
    linkcolor={blue!50!black},
    citecolor={blue!50!black},
    urlcolor={blue!80!black}
}

\usepackage{amssymb}

\usepackage[numbers]{natbib}

\usepackage[shortcuts,acronym,nonumberlist]{glossaries}
\makeglossaries %

\newacronym{RMSE}{RMSE}{Root Mean Squared Error}
\newacronym{PINN}{PINN}{Physics-Informed Neural Network}
\newacronym{ANN}{ANN}{artificial neural network}
\newacronym{DNN}{DNN}{deep neural network}
\newacronym{TBP}{TBP}{three-body problem}
\newacronym{ML}{ML}{machine learning}
\newacronym{HNN}{HNN}{hamiltonian neural network}
\newacronym{PI}{PI}{physics-informed}

\begin{document}

        \twocolumn[
        \begin{@twocolumnfalse}

            \vspace{0pt}
            \begin{center}
                \selectfont\fontsize{10}{0}\selectfont IAC-24,C1,IPB,13,x87864

                \vspace{15pt}
                \textbf{Advancing Solutions for the Three-Body Problem Through Physics-Informed Neural Networks
                }

                \vspace{10pt}
                \textbf{
                    \selectfont\fontsize{10}{0}\selectfont Manuel Santos Pereira~\textsuperscript{a*}, Luís Tripa~\textsuperscript{a}, Nélson Lima\textsuperscript{b}, Francisco Caldas\textsuperscript{a}, Cláudia~Soares~\textsuperscript{a}
                }
            \end{center}

            \vspace{-10pt}
            \begin{flushleft}

                \textsuperscript{a}\textit{
                    \fontfamily{ptm}\selectfont\fontsize{10}{12}\selectfont NOVA School of Science and Technology, Caparica, Portugal},\\
                \small{\underline{mdr.pereira@campus.fct.unl.pt},
                    \underline{l.tripa@campus.fct.unl.pt},
                    \underline{f.caldas@campus.fct.unl.pt}, \underline{claudia.soares@fct.unl.pt}}
                \\
                \textsuperscript{b}\textit{
                    \fontfamily{ptm}\selectfont\fontsize{10}{12}\selectfont Universidade de Coimbra, Coimbra, Portugal},\\
                \small{\underline{ uc2003107901@student.uc.pt
                }}
                \\
                \textsuperscript{*}\fontfamily{ptm}\selectfont\fontsize{10}{12}\selectfont Corresponding Author
            \end{flushleft}

            \begin{abstract}                
                First formulated by Sir Isaac Newton in his work ``\textit{Philosophiæ Naturalis Principia Mathematica}'' the concept of the Three-Body Problem was put forth as a study of the motion of the three celestial bodies within the Earth-Sun-Moon system. In a generalized definition, it delves into the prediction of motion for an isolated system composed of three point masses freely interacting under Newton's law of universal attraction. This proves to be analogous to a multitude of interactions between celestial bodies, and thus, the problem finds great applicability within the studies of celestial mechanics. Despite its importance and numerous attempts by renowned physicists to solve it throughout the last three centuries, no general closed-form solutions have been reached due to its inherently chaotic nature for most initial conditions. Current state-of-the-art solutions are based on two approaches, either numerical high-precision integration or machine learning-based. Notwithstanding the amazing breakthroughs of neural networks, these present a significant limitation, which is their ignorance of any prior knowledge of the problem and chaotic systems presented. Thus, in this work, we propose a novel method that utilizes Physics-Informed Neural Networks (PINNs). These deep neural networks are able to incorporate any prior system knowledge expressible as an Ordinary Differential Equation (ODE) into their learning processes as a regularizing agent. Our model was trained on a dataset composed of 30,000 simulations made with recourse to a high-precision n-body integrator. The simulated data belonged to systems with three unitary masses and initial velocity equal to zero, corresponding to a restricted form of the problem. Our findings showcase that PINNs surpass current state-of-the-art machine learning methods with comparable prediction quality. Despite a better prediction quality, the usability of numerical integrators suffers due to their prohibitively high computational cost. These findings confirm that PINNs are both effective and time-efficient open-form solvers of the Three-Body Problem that capitalize on the extensive knowledge we hold of classical mechanics.

                \vspace{0.25em}
            \end{abstract}

        \end{@twocolumnfalse}
    ]

    \section*{Acronyms/Abbreviations}
    
    \vspace{3mm}

    \begin{description}
        \item ANN: Artificial Neural Network
        \item DNN: Deep Neural Network
        \item HNN: Hamiltonian Neural Network
        \item PI: Physics-Informed
        \item PINN: Physics-Informed Neural Network
        \item RMSE: Root Mean Square Error
        \item TBP: Three-Body Problem
    \end{description}

    \section{Introduction}

Initially formulated by Sir Isaac Newton in his 1687 work, \emph{"Pricipia Mathematica"} \cite{Newton1687}, the \gls{TBP} consists in the study of motion of three celestial bodies modelled by Newtonian mechanics. The initial studies by Newton focused on the motion of the Sun, Earth and Moon, expressed as a system, and attempted to infer their orbits in some meaningful fashion. From that point onward, the three-body problem became a central point in mathematical physics, drawing the attention of renowned physicists, such as \citet{Euler1767} and \citet{Lagrange1772}. Several of the attempts to solve the problem drew new perspectives forward, as is the case of the work of \citet{Poincare1893}, who demonstrated that the problem could not be solved by analytical means due to the chaotic nature of the system. More than 200 years after its initial proposal, \citet{Sundman1909} discovered a solution in the form of an infinite power series.

To formally define the three-body problem under the framework of Newtonian mechanics, let $\mathbf{\rho}_i \equiv (x_i, y_i, z_i)$ be a three-dimensional spatial vector which designates the location of the $i$-th body with respect to a given initial frame. As the system is assumed to be isolated, the sole force acting upon the bodies is gravitational, and we get the following system of equations:
\begin{gather}
    \begin{split}
        \mathbf{\ddot{\rho}}_1 = -m_2 \frac{\mathbf{\rho}_1 - \mathbf{\rho}_2}{\|\mathbf{\rho}_1 - \mathbf{\rho}_2\|^3} - m_3 \frac{\mathbf{\rho}_1 - \mathbf{\rho}_3}{\|\mathbf{\rho}_1 - \mathbf{\rho}_3\|^3},\\
        \mathbf{\ddot{\rho}}_2 = -m_1 \frac{\mathbf{\rho}_2 - \mathbf{\rho}_1}{\|\mathbf{\rho}_2 - \mathbf{\rho}_1\|^3} - m_3 \frac{\mathbf{\rho}_2 - \mathbf{\rho}_3}{\|\mathbf{\rho}_2 - \mathbf{\rho}_3\|^3},\\
        \mathbf{\ddot{\rho}}_3 = -m_1 \frac{\mathbf{\rho}_3 - \mathbf{\rho}_1}{\|\mathbf{\rho}_3 - \mathbf{\rho}_1\|^3} - m_2 \frac{\mathbf{\rho}_3 - \mathbf{\rho}_2}{\|\mathbf{\rho}_3 - \mathbf{\rho}_2\|^3},
    \end{split}
    \label{eq:tbp-system}
\end{gather}
where $m_n$ represents the mass of the n-th body \cite{Broucke1973}, $\ddot{\rho}$ represents the second derivative of $\rho$ \emph{w.r.t.} time \cite{Broucke1973}, and $\| \mathbf{v} \|$ represents the Euclidean norm of vector $\mathbf{v}$. Additionally, we may gather relevant information when expressing the \gls{TBP} under the Lagrangian formalism, defining the problem as: 
\begin{align}
    \begin{split}
        \mathcal{L} = &\frac{1}{2}\left[m_1\mathbf{\dot{\rho}}_1 + m_2\mathbf{\dot{\rho}}_2 + m_3\mathbf{\dot{\rho}}_3 \right]\\ &+ \left[ \frac{m_2m_3}{\|\mathbf{\rho}_3 - \mathbf{\rho}_2\|} + \frac{m_1m_3}{\|\mathbf{\rho}_3 - \mathbf{\rho}_1\|} + \frac{m_1m_2}{\|\mathbf{\rho}_2 - \mathbf{\rho}_1\|}\right].
    \end{split}
\end{align}
under this formalism, phenomena are described using the energies of the system. Within this specific system, two energies are defined, the total kinetic energy - represented by the first term of the Lagrangian ($\mathcal{L}$) - and the gravitational potential energy of the system - displayed by the second term.

\subsection{Existing (periodic) solutions}

Due to the chaotic nature of the three-body problem, and the non-existence of closed-form analytical solutions, periodic solutions to the three-body problem ``\dots are, so to say, the only opening through which we can try to penetrate in a place which, up to now, was supposed to be inaccessible.'' \cite{Poincare1893}. 

Until the begining of the 21st century, only three families of periodic orbits had been discovered \cite{Suvakov2013}: 
\begin{enumerate}
    \item Euler-Lagrange
    \item Broucke-Henon-Hadjidemetriou (BHH)
    \item Figure-Eight \cite{Moore1993}
\end{enumerate} 

The most recent in the discovery of families of periodic solutions came by \citet{Li2017} in 2017, who reported on the finding of more than 600 new families of periodic solutions of the equal-mass problem $(m_1 = m_2 = m_3)$, and the discovery of 1349 new families of two equal-massed body systems two years later in 2019 \cite{Liao2019}.

These discoveries have greatly expanded our understanding of the complex dynamics of the three-body problem and the diversity of periodic solutions.

\subsection{Applicability of the three-body problem}

The integrated study of machine learning and the three-body problem can offer a significant amount of insight into both fields of study.

Firstly, the three-body problem may be seen as a case study on the application of machine learning architectures to chaotic systems. Given that many real-world systems and phenomena, even in something as conceptually simple as a passive double pendulum \cite{Levien1993}, demonstrate a chaotic nature, the insights gained here could potentially improve the performance of future models on such phenomena.

On the other hand, by gaining insight into the three-body problem we might gain knowledge of the complex interactions within astronomical and celestial dynamics, such as for example within study on the formation of close binaries.
Lastly, better understanding of celestial mechanics and the behaviour of bodies under the three-body framework may allow improved studies on the existence of exoplanets by determining the effect of three-body interactions against linear predictions of transit time \cite{Musielak2014}. 

\subsection{Machine learning and the physical sciences}

The intersection of machine learning and physics represents a rapidly developing body of scientific research. Which, despite the relative infancy of machine learning (ML) when compared to the study of physics, has already played a crucial role in recent discoveries and research. Applications span a wide spectrum, ranging from classical mechanics to particle physics and quantum systems \cite{Carleo2019}, demonstrating the adaptability of this approach. 

Conversely, physics also has the ability to contribute to the development and training of neural network models. Due to the conjunction of high data requirements of neural networks \cite{Bishop2023} with the exceptionally high data acquisition cost of many physical fields of study, training of neural networks becomes unfeasible with real-world data. By taking advantage of physical modelling, synthetic data can be simulated at a fraction of the cost, enabling previously impossible training. 

Lastly, trends show a year-after-year increase in the development of physics-inspired methods for machine learning, with many influential works being proposed since 2018 \cite{Raissi2018, Pathak2018, Cranmer2020, Greydanus2019}.

\subsection{Contributions}

This work addresses the incorporation of the vast amount of physical knowledge we hold on classical mechanics, specifically of the well-defined three-body problem, into machine learning methodologies. Our primary aim is to demonstrate the effectiveness of physics-informed neural networks (PINNs) in enhancing the learning capabilities and inference accuracy comparatively to standard neural network models.

To this effect we provide a comprehensive comparison between our \gls{PINN} methodology and standard deep neural networks, building upon and extending the work of \citet{Breen2020}. This comparison offers insights into the advantage of the inclusion of physically-informed losses in terms of accuracy and generalization.

Through these contributions, we aim to add to the growing body of research regarding the integration of machine learning techniques to the physical sciences, additionally showcasing the viability of physical priors as regularization terms in neural networks.

    \section{Neural Networks and Physics Priors}
\label{sec:nn-and-pinn}

\begin{figure}[!ht]
    \centering
    \includegraphics[width=\linewidth]{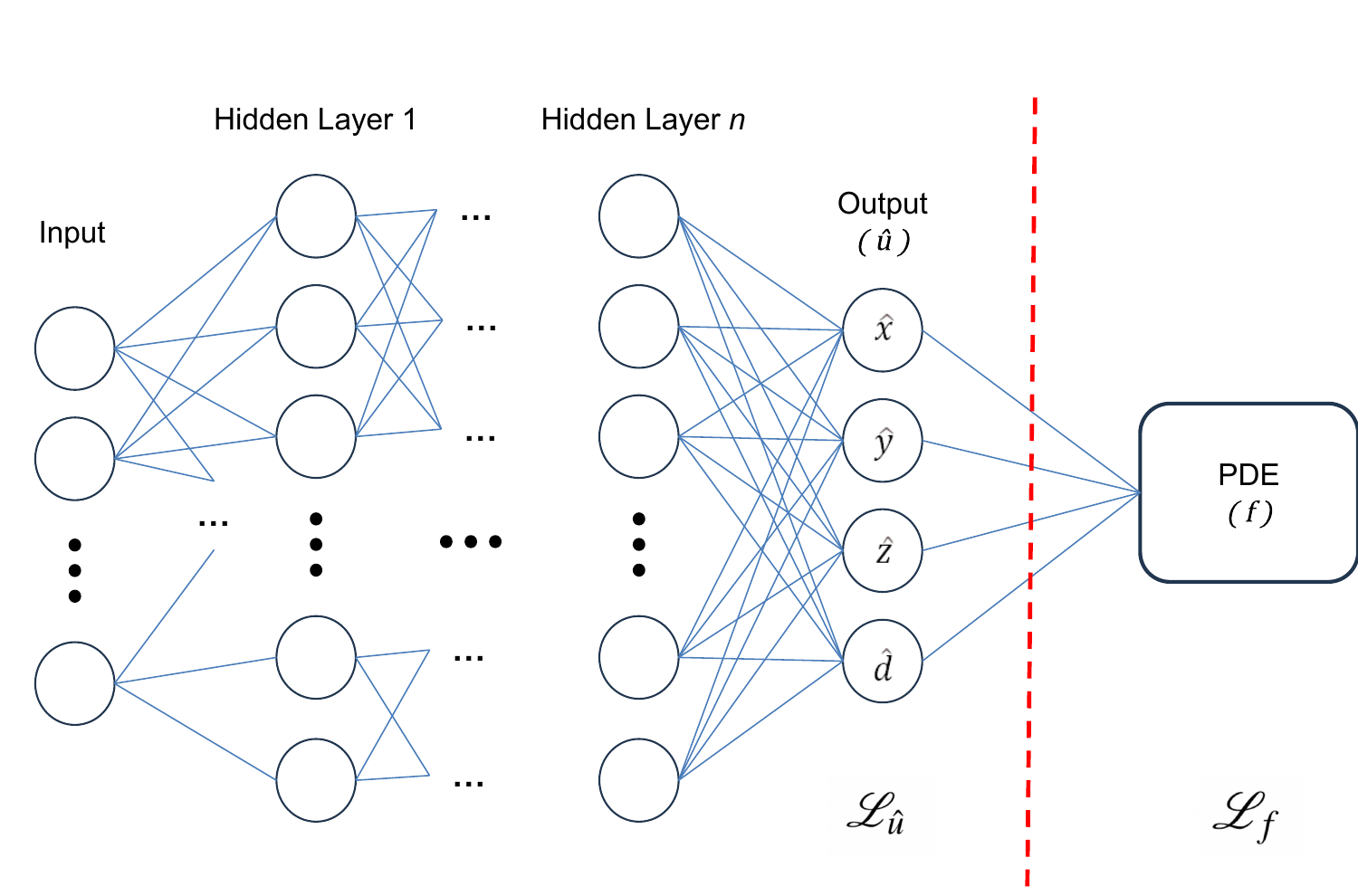}
    \caption{Simplified schemata of a \gls{PINN}, illustrating the difference to classical \glspl{ANN}. The section to the left of the red vertical line depicts a standard \gls{ANN}.} 
    \label{fig:pinn}
\end{figure}

\Glspl{ANN}, visualized in Figure \ref{fig:pinn}, represent a specific subfield of machine learning models, and can be formally defined as continuous parametric functions constructed through the layered application of linear and nonlinear operators initially thought to emulate the simplified operation of biological neurons \cite{Fukushima1975}. However, scientific consensus has shifted from this biological view to a purely mathematical object. 

These computational units are fit to the data through an optimization process that attempts to minimize the discrepancy between the outputs of the function and a desired solution, termed ``ground truth''; this discrepancy is termed ``loss'', and is denoted as $\mathcal{L}$ \cite{Bishop2023}. Such minimization is performed through a gradient descent family of algorithms, and thus requires sub-differentiability.

\subsection{Physics-Informed Neural Networks}

First proposed by \citet{Lagaris1998} in 1998, and further developed in 2019 by \citet{Raissi2019},
\glspl{PINN} are a class of machine- and deep learning methods designed to solve forward and inverse physical problems. They represent an improved solver of physics problems due to the inclusion in the model of the vast prior mathematical and physical knowledge we may hold of the systems under study. This inclusion of priors not only allows for a reduction in the necessary data for training, which in the context of physical problems that usually present prohibitive data acquisition costs, allows for an improved performance of the network, but also forces results to assume physically realistic values.

Non-linear parametric partial differential equations (PDEs) are considered, defined as
\begin{equation}
    \dot{u} + \mathcal{N}[u, \lambda] = 0, \; x \in \mathbb{R}^d, \; t \in [0, T],
\end{equation}
where $u(t, x)$ denotes the latent solution and $N [\cdot, \lambda]$ is a nonlinear operator parametrized by $\lambda$. 

As a primary use for \glspl{PINN}, the author proposes the data-driven finding to solutions of partial differential equations. Within the scope of this framework, a general-form partial differential equation is defined as:
\begin{gather}
    \dot{u} + \mathcal{N}[u; \lambda] = 0, \; x \in \mathbb{R}^d,\; t \in [0, T],\\
    f := \dot{u} + \mathcal{N}[u; \lambda]. \label{eq:pinn-func}
\end{gather}
By modelling $u(t,x)$ through a learned function, and imposing the assumptions stated through Eq.~\eqref{eq:pinn-func} we have effectively created a physics-informed neural network. The learning process of the network is performed through a family of gradient descent methods, optimizing the following loss function:
\begin{equation}
    \mathcal{L} = \mathcal{L}_\mathrm{MSE}^{(u)} + \mathcal{L}_\mathrm{MSE}^{(f)},
\end{equation}
composed of the sum of two mean-square-error (MSE) losses, defined as
\begin{align}
    \mathcal{L}_\mathrm{MSE}^{(u)} &= \frac{1}{N_u} \| u(t^i_u, x^i_u) - u^i \|_2^2,\label{eq:pinn-Lu}\\
    \mathcal{L}_\mathrm{MSE}^{(f)} &= \frac{1}{N_f} \| f(t^i_f, x^i_f) \|_2^2,\label{eq:pinn-Lf}
\end{align}
wherein  we define the set $\{t^i_u, x^i_u, u^i\}^{N_u}_{i=1}$ as the initial data on $u(t,x)$ and
$\{t^i_f, x^i_f\}^{N_f}_{i = 1}$ define collocation points for the PDE residual, $f(t,x)$. As such, we may interpret the first term of the loss as penalizing discrepancies between predicted and ground truth values, and the second term guaranteeing that solutions satisfy the PDE.

    \section{Literature Review}

Due to its critical role in understanding celestical dynamics, the three-body problem has been the subject of extensive research since its definition. Over time, numerous methodologies have been put forth as solutions to the problem, each offering differing degrees of complexity and success.

\subsection{Numerical Methods}

In part due to the lack of closed-form solutions to the three-body problem \cite{Poincare1893,Musielak2014}, numerical methods have proven to be indispensible tools for generating solutions. However, due to their autoregressive nature, numerical solvers suffer from exponential growth of errors \cite{Goodman1993, Boekholt2014} as numerical errors are accumulated and propagated over time.

Some notable examples of numerical methods include those implemented by \citet{Wisdom1992}, which utilizes symplectic integrators to preserve the Hamiltonian structure of the system over longer time scales; \citet{Press1992} who utilize the Runge-Kutta \cite{Runge1895} method to approximate solutions to the Partial Differential Equations describing the system; and lastly \citet{Boekholt2014}, implementing BRUTUS, a numerical integrator based on the Bulrisch-Stoer algorithm \cite{Bulirsch1966}.

Generation of data within this work was performed using the BRUTUS algorithm \cite{Boekholt2014}, and is described further in Section \ref{subsec:data}.

\subsection{Machine Learning-based Methods}

Analysis of relevant literature presented the study by \citet{Breen2020} as one of the most compelling works on the integration of machine learning and the three-body problem. The author has implemented a vanilla feed-forward \gls{DNN}, which can be seen as the left side of Figure \ref{fig:pinn}, and limited the study to a fixed time interval of $t \leq 10$ time units. For this task an architecture composed of 10 hidden layers at a width of 128 units per layer was designed, and each layer was interspersed with rectified linear unit activations (ReLU), defined by \(h(x) = \max(0, x)\). Training was performed using the ADAM \cite{Kingma2017} gradient descent algorithm over 10000 batches of 5000 datapoints each.

Further works regarding the application of machine learning to the three-body problem by \citet{Kumar2021} and \citet{Mi2021} explore Reservoir Computing \cite{Pathak2018} and \Glspl{HNN} \cite{Greydanus2019} as potential promising alternatives to vanilla neural networks, however they do not include comprehensive results to validate these approaches.

Lastly, \citet{Choudhary2020} present the application of \Glspl{HNN} \cite{Greydanus2019} to chaotic dynamical systems under the Hénon-Heiles model, and prove that \Glspl{HNN} are able to operate within chaotic regimes, while conserving the total energy of the system.

    \section{Methods and Data}
    \label{sec:methods-and-data}

    \subsection{Dataset}
    \label{subsec:data}
    
\begin{figure}[h]
        \centering
        \includegraphics[width=\linewidth]{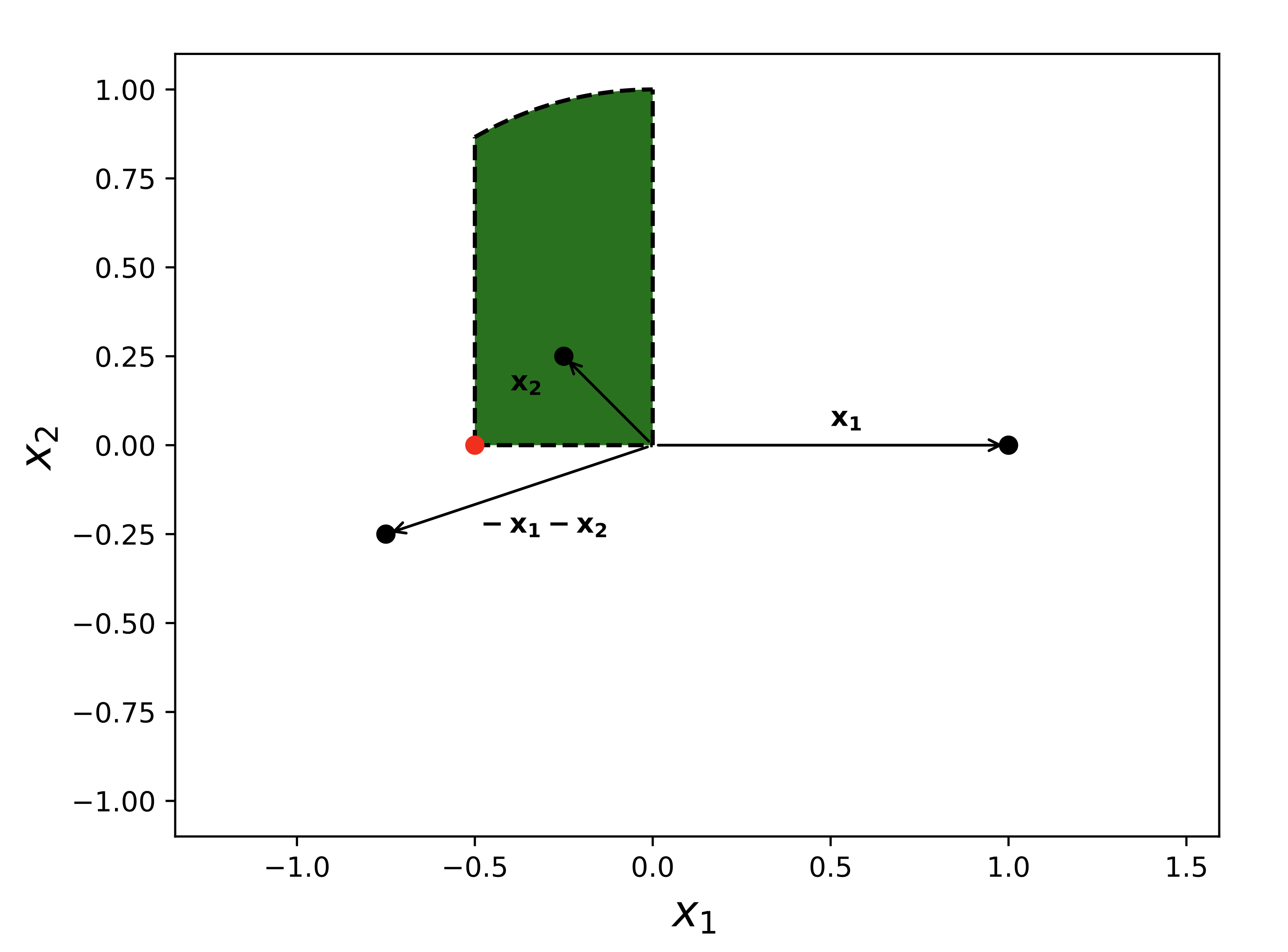}
        \caption{Illustration of the particle's reference plane, where the green area represents all the valid positions where $\mathbf{p}_2$ can be initially placed. The red point represents the singularity, a point where $\mathbf{p}_2$'s and $\mathbf{p}_3$'s positions coincide. Figure adapted from \cite{Breen2020}.}
        \label{fig:reference_plane}
    \end{figure}

    To train the model effectively and enable accurate generalization to unseen data, it must be trained on a dataset that is both representative and diverse. For this work, a dataset was simulated by following the methodology of \citet{Breen2020}. In that work, the authors fix the position of particle 1 ($\mathbf{p}_1$), while the position of particle 2 ($\mathbf{p}_2$) is generated within predetermined constraints. Finally, the position of particle 3 ($\mathbf{p}_3$) is obtained through a linear combination of the positions of the other two particles. These constraints ensure that the system is initialized with its center of mass at the origin, which is a common simplification in the study of the three-body problem. This methodology generates initial positions similar to those visualized in Figure \ref{fig:reference_plane}.
    
    Therefore, it is straightforward to obtain the positions $\mathbf{p}_1$ and $\mathbf{p}_3$ of the first and third bodies, where any $\mathbf{p}_n \equiv (x_n, y_n = 0, z_n)$ is a three-dimensional spatial vector. However, $\mathbf{p}_2$ must be randomly placed within a defined area while ensuring a uniform sampling of this space to maintain diversity within the dataset. To achieve this, we propose Algorithm \ref{alg:init-cond} to generate meaningful initial positions for the bodies.
    
    \begin{algorithm}[h]
        \caption{Generation of initial conditions for simulation.}
        \label{alg:init-cond}
        \begin{algorithmic}[1]
            \State Sample $\theta$ uniformly from $[0, \frac{\pi}{2}]$.
            \item[]
            \State Create $\mathbf{p} \equiv (x,y,z)$ with the following components:
    
            $x \gets -\min(0.5, \cos(\theta))$
    
            $y \gets 0$ 
    
            $z \gets \sin(\theta)$
            \item[]
            \State Sample $s$ uniformly from $[0, 1]$.
            \State Multiply $\mathbf{p}$ by $s$ to calculate the final position of $p_2$.
        \end{algorithmic}
    \end{algorithm}
    
    Following the generation of the initial positions $\mathbf{p}_{1,2,3}$, the BRUTUS integrator \cite{Boekholt2014} was used to simulate the motions of the three particles for a total of $t = 10$ time units, with a step size of $\Delta t = 0.0390625$, which corresponds to a total of $256$ time steps per simulation. This integration technique is parametrized by two hyperparameters: the word length $L_w$, which corresponds to the bytes used to represent numerical values, and the tolerance $\varepsilon$, which represents the maximum accepted deviation between individual integrators. Following the work of \citet{Breen2020}, these were set to $L_w = 88$ and $\varepsilon = 1.0 \times 10^{-10}$.
    
    As a simplification of the problem, we assumed a unitless system where both the gravitational constant $G$ and the particle masses $m_1 = m_2 = m_3$ were set to $1$. The initial velocities of each particle were assumed to be $0$ across all axes. Additionally, the planar formulation of the three-body problem was adopted, following the methodology of \citet{Breen2020}, which involves assuming constant values for both position and velocity along the $y$-axis. Lastly, the bodies were assumed to be perfectly represented by point particles, ensuring that the probability of collisions is infinitesimally small.
    
\begin{figure*}
        \centering
        \begin{subfigure}[t]{0.45\textwidth} %
            \centering
            \includegraphics[width=\textwidth]{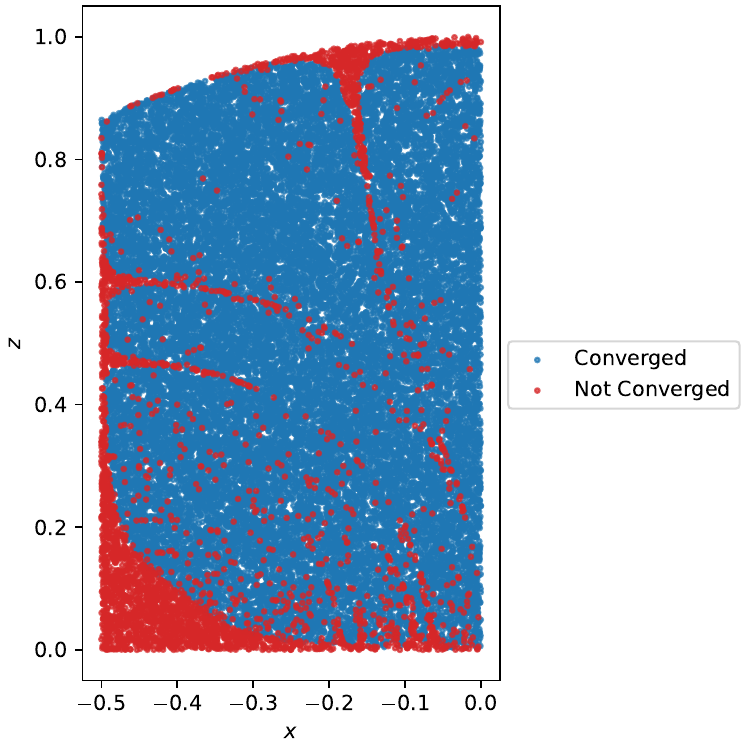} %
            \caption{Illustration of the relation between the location of particle $2$ and the convergence of the numerical integrator. Points in blue represent converged simulations, while those in red represent simulations that did not converge.}
            \label{fig:x2-vs-conv}
        \end{subfigure}
        \hfill
        \begin{subfigure}[t]{0.45\textwidth} %
            \centering
            \includegraphics[width=\textwidth]{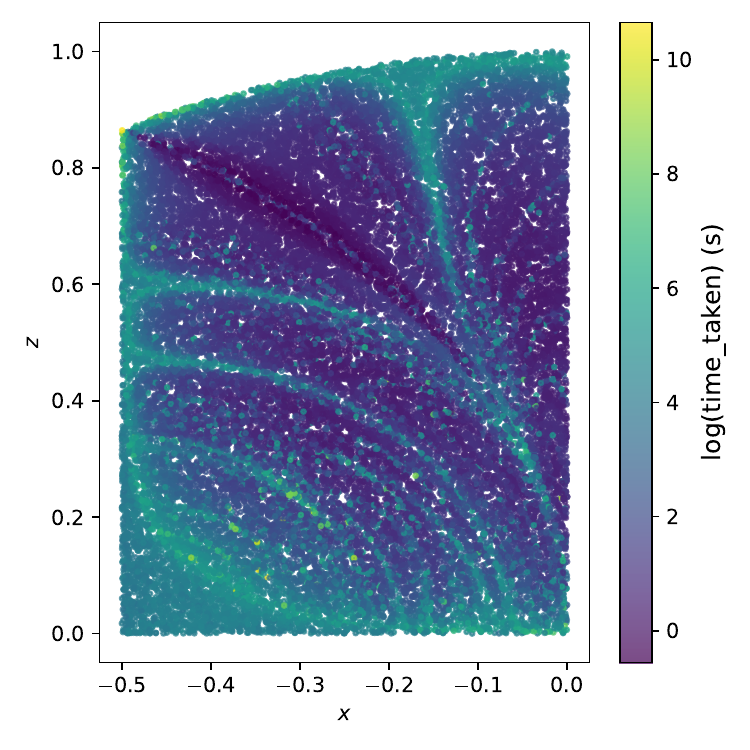} %
            \caption{Illustration of the relation between the location of particle $2$ and the time to necessary to calculate the full timescale of the simulation.}
            \label{fig:x2-vs-time-taken}
        \end{subfigure}
        \caption{Visualization of the dependence of the BRUTUS three-body simulations on the choice of initial position for particle $2$.}
        \label{fig:x2-vs-multiple}
    \end{figure*}

    In total, $30000$ simulations unique simulations where generated, of which around $20\%$ did not converge numerically. This divergence occurred mostly when $\mathbf{p}_2$'s position was near the singularity point, where particles $\mathbf{p}_2$ and $\mathbf{p}_3$ are close to each other. Figure \ref{fig:x2-vs-conv} visualizes the distribution of $\mathbf{p}_2$'s position for all simulations and whether they converged.
    
    \FloatBarrier

\subsection{Architecture}
\label{subsec:architecture}

\begin{figure}[h!]
        \centering
        \includegraphics[width=\linewidth]{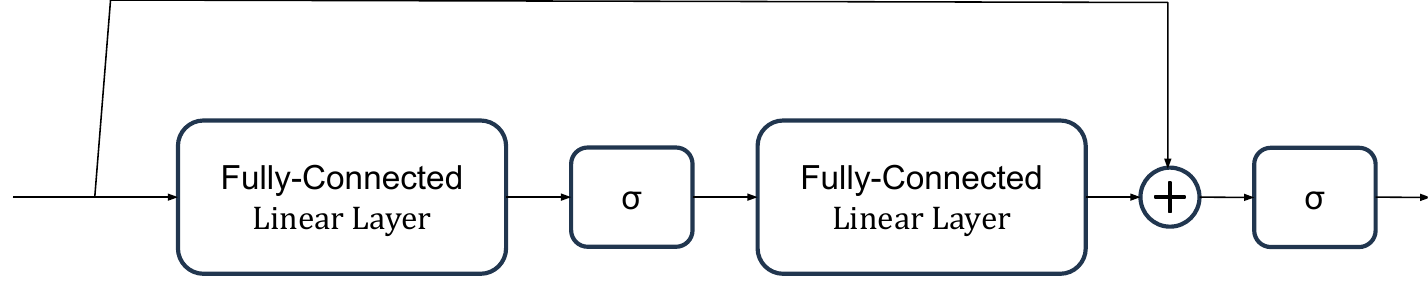}
        \caption{Residual block, where $\sigma$ represents any nonlinear activation function and a Fully Connected Linear Layer is a layer of neurons where each is connected to each of the inputs and outputs. A ResNet is composed of successive applications of this block.}
        \label{fig:resnet-block}
    \end{figure}

    Due to the added constraints introduced by the physics-informed term of the loss, \glspl{PINN} must be more expressive than their counterparts. This increase in expressiveness is achieved by deepening or expanding its width, effectively increasing the number of parameters. In addition to this requirement, the increased complexity of the loss landscape makes PINNs susceptible to the ``exploding gradients problem'', which causes updates to the network to become highly volatile, and eventually destabilizes the training process irreversably \cite{Sharma2023,Bishop2023}. So as to improve gradient flow and possibly avoid this issue, we additionally implemented the ResNet \cite{He2016} architecture. This architecture is similar to a standard feed-forward network, with the addition of residual connections which take the output of layer $l$, and add it to the output of layer $l+n$, visualized in Figure \ref{fig:resnet-block}. 
    
    Given the comparative nature of this study, we have built on the methodology outlined by \citet{Breen2020}. The original author proposed a standard \emph{feed-forward} deep neural network, this is, a network where no closed directed cycles are formed in the neuron graph \cite{Bishop2023}.

    Each architecture was additionally formulated under both a non-autoregressive and an autoregressive setup. Firstly, under the non-autoregressive formulation, the model takes as input a 7-vector containing the initial positions of bodies and a time $t_n$, and subsequently predicts a 12-vector representing the state of the system at $t_n$. This output vector is thus composed of both the body positions and velocities. Contrastively, the autoregressive setup takes as input the state of the system at any time $t_0$ and a timestep $\Delta t$, and is tasked with predicting the state of the system at timestep $t = t_0 + \Delta t$.

    As baseline, we have implemented a feed-forward deep neural network composed of 12 layers, each of which may be separated into three categories based on their task: input, hidden and output. A single input layer is used, composed of 7 or 13 units ("neurons"), for non-autoregressive and autoregressive formulations respectively. A total of 10 hidden layers were used, each at a width of 128 units \cite{Breen2020}. Lastly, a single output layer was used composed of 12 units, each responsible for outputting the inferred value associated to an individual variable of the system. Each layer is interspersed with a rectified linear unit (ReLU), a nonlinearity that allows the model to express non-linear functions \cite{Bishop2023}.

    Choices of design for the PINN are discussed in the results section, specifically in Table \ref{tab:hyperparameter-optimization}. 

\subsection{(Physics-Informed) Loss}
\label{subsec:loss}

    As a comparative baseline, we take the loss used by \citet{Breen2020}, this being a mean-absolute-error (MAE) loss, defined by:
    \begin{align}
        \begin{split}
            \mathcal{L}_{\mathrm{MAE}} &= \frac{1}{N_u} \| u(\mathbf{t}_u, \mathbf{x}_u) - \mathbf{u} \|_1\\
            &= \frac{1}{N_u}\sum_{i=1}^{n} |u(t^i_u, x^i_u) - u^i| 
        \end{split}
        \label{eq:pinn-Lu-mae}
    \end{align}
   where we define the set $\{t^i_u, x^i_u, u^i\}^{N_u}_{i=1}$ as the initial data on $u(t,x)$. The MAE loss is thus equivalent to Eq.~\eqref{eq:pinn-Lu}, and serves the purpose of punishing discrepancies between the ground truth and model predictions.

   As priorly mentioned in Section \ref{sec:nn-and-pinn}, physics-informed losses are composed of not only the data discrepancy loss, but also a (partial) differential equation collocation loss that punishes model predictions which do not correspond to the laws of the system, denoted  $\mathcal{L}^{(f)}$, as in Eq. \eqref{eq:pinn-Lf}. 
   
   The total loss of the model may then be defined as
   \begin{equation}
    \mathcal{L} = \mathcal{L}_{\mathrm{MAE}}^{(u)} + \alpha \mathcal{L}^{(f)}
   \end{equation}
   where $\alpha$ is a weighing coefficient attached to the physics-informed loss term. Choice of alpha is discussed in the results section, specifically in Table \ref{tab:hyperparameter-optimization}.

    \subsection{Training Regime}

    According to best practices, the total dataset was partitioned into two sub-sets, a training dataset composed of 95$\%$ of the data used to update the parameters of the model, and a validation set used to verify its ability to generalize \cite{Bishop2023}.

    A model is said to overfit when evaluation on training and validation datasets diverges, favouring the training set, which equates to the model losing generalization capabilities. In order to avoid overfitting, early stopping was implemented, stopping the training process after 10 epochs after the last improvement over the validation set \cite{Prechelt1998}.

    Models were trained using the ADAM \cite{Kingma2017} gradient descent method for a maximum of 500 epochs. In addition, due to the complexity of the loss landscape, learning rate was scheduled to reduce by a factor of $0.7$ in case of a plateau of validation performance lasting 5 epochs.

    Further choices of training regime are further discussed in the results section, specifically under Table \ref{tab:hyperparameter-optimization}.

\section{Hyperparameter optimization}
\label{subsec:ablation-study}

In addition to the model parameters, there exist so-termed \emph{hyperparameters}, which are not affected during the minimization (learning) process. Some notable examples are network depth, network width, learning rate and batch size \cite{Bishop2023}. These parameters can have profound effects on learning and must be adequatly optimized upon.

To this effect, a manual grid search was performed over an arbitrary parameter range, followed by a local search in the neighbourhood of the best-performing value. The results of this search will be discussed in the following subsections, and are summarized in Table \ref{tab:hyperparameter-optimization}.

\renewcommand{\arraystretch}{1.3}

\begin{table*}[ht]
    \caption[Hyperparameter Optimization]{Results of hyperparameter optimization. Elements marked in bold indicate best performance within the group.}
    \label{tab:hyperparameter-optimization}
    \centering
    \begin{tabularx}{\linewidth}{|>{\centering\arraybackslash}p{0.25\linewidth}!{\vrule width 1.5pt}C|C|C|C|}
        \hline
        \textbf{Architecture} & \multicolumn{2}{c|}{\textbf{Non-Autoregressive}} & \multicolumn{2}{c|}{Autoregressive} \\
        \hline
        \textbf{Architecture (2)} & \multicolumn{2}{c|}{Standard DNN} & \multicolumn{2}{c|}{\textbf{ResNet DNN}} \\
        \hline
        \textbf{Network Depth} & 6 & 9 & \textbf{12} & 15 \\
        \hline
        \textbf{Network Width} & 64 & 128 & \textbf{256} & 512 \\
        \hline
        \textbf{Activation Unit} & GELU & \textbf{ReLU} & Tanh & Leaky ReLU \\
        \hline
        \textbf{Learning Rate} & 5e-4 & \textbf{7.5e-4} & 1e-3 & 5e-3 \\
        \hline
        \textbf{Gradient Norm Clipping} & \textbf{1} & \textbf{3} & \textbf{5} & 7 \\
        \hline
        \textbf{Weight Decay} & \multicolumn{2}{c|}{\textbf{1e-5}} & \multicolumn{2}{c|}{No weight decay} \\
        \hline
        $\mathbf{\alpha}$ \textbf{scheduler} & Constant & Warmup & \textbf{Linear} & Exponential \\
        \hline
        \textbf{Initial} $\mathbf{\alpha}$ & \textbf{0.001} & 0.01 & 1 & 10 \\
        \hline
        \textbf{Final} $\mathbf{\alpha}$ & 0.5 & \textbf{0.75} & 1 & 1.25 \\
        \hline
        \textbf{Clamp} $\mathcal{L}_f$ & 1e-6 & 1e-4 & 1e-2 & \textbf{No clamp} \\
        \hline
    \end{tabularx}
\end{table*}

\subsection{Model Formulation}

As previously discussed in Subsection \ref{subsec:architecture}, models were formulated under a non-autoregressive and autoregressive format. We hypothesized that autoregressive architectures would exhibit lower errors at smaller intervals, as their predictions are conditioned on prior outputs over these short time steps. However, similar to numerical integrators, that errors would accumulate and be propagated over successive iterations, leading to an exponential growth. In contrast, we believed non-autoregressive architectures would exhibit larger training errors at lower iterations, as training is not focused on these smaller timesteps, but be resilient to larger intervals by avoiding the accumulation of errors. 

\begin{figure}[H]
    \centering
    \includegraphics[width=\linewidth]{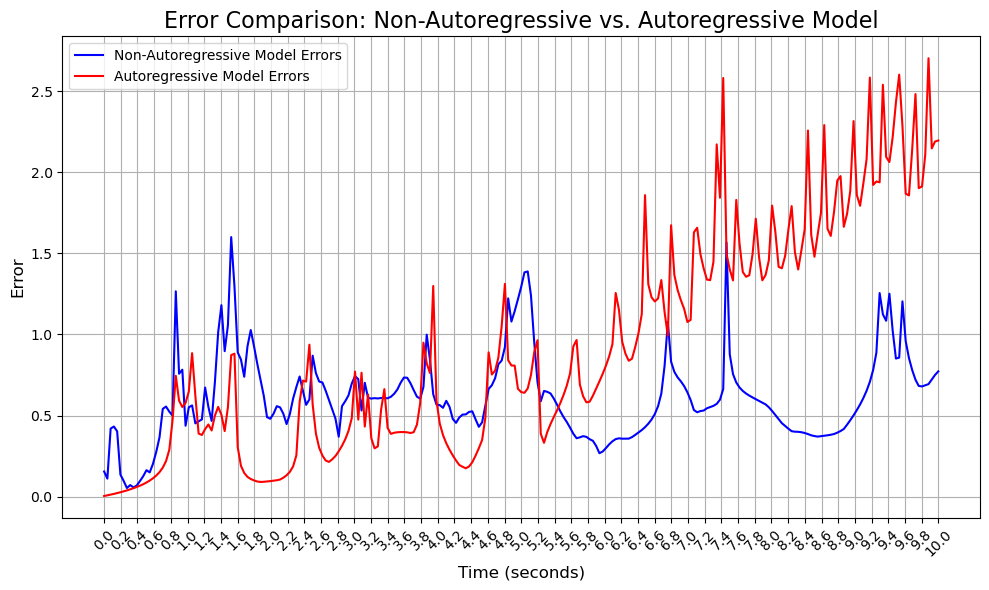}
    \caption{Error comparison over time between autoregressive and non-autoregressive models, highlighting the exponential error growth of autoregressive models.}
    \label{fig:error-t-regchoice}
\end{figure}

Results confirm our initial hypotheses, indicating that non-autoregressive models perform worse than their counterparts at lower intervals, and also demonstrate the exponential trend of errror accumulation of autoregressive models. 

\subsection{Alpha coefficient}

Previously discussed under Subsection \ref{subsec:loss}, the physics-informed loss term is weighted through a hyperparameter coefficient, $\alpha$. Results indicated that lower values $(\alpha \approx 0.001)$ provided good performance for lower epochs, while benefits tended to stagnate at 100 epochs. Contrastively, higher alpha parameters made training significantly volatile in earlier epochs, but benefited later training stages.

Given these results, a scheduler was implemented to dynamically attribute $\alpha$ values, including a warmup stage wherein the value was set to zero $(\alpha_0 = 0)$ until an arbitrary number of epochs had passed, after which $\alpha$ was sharply increased to a desired $\alpha_{\mathrm{max}}$ to increase the influence of the physics-informed loss term. This approach significantly improved training of earlier epochs, as the complexity of the physics-informed loss term was circumvented; however, the sharp increase from $\alpha_0$ to $\alpha_{max}$ caused gradients to explode and training to degenerate.

\begin{figure}[H]
    \centering
    \includegraphics[width=\linewidth]{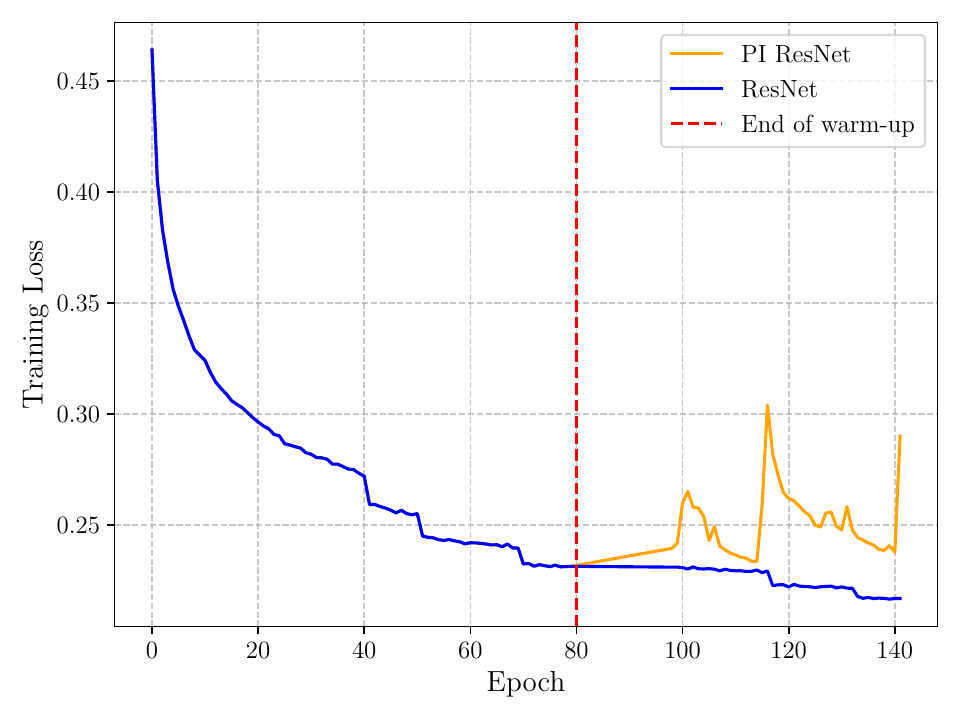}
    \caption{Comparison of training loss between standard DNN and PINN with sharp $\alpha$ increase after warmup.}
    \label{fig:warmup-no-ramp}
\end{figure}

As a response to this issue, a ramping stage was added to the warmup scheduler, which would gradually increase $\alpha_0$ to match $\alpha_{max}$ in a linear or exponential fashion. Contrary to expectations, despite the gradual introduction of $\alpha$, training could not cope and would invariably degenerate.

These findings seem to suggest minima proposed by the standard data localization-focused MAE loss are incompatible with minima of the physics-informed loss term. Consequently, we propose training to be performed under supervision of both loss terms, with a gradual increase of $\alpha$ to maintain significance of the physics-informed loss term. 

\begin{figure}[H]
    \centering
    \includegraphics[width=\linewidth]{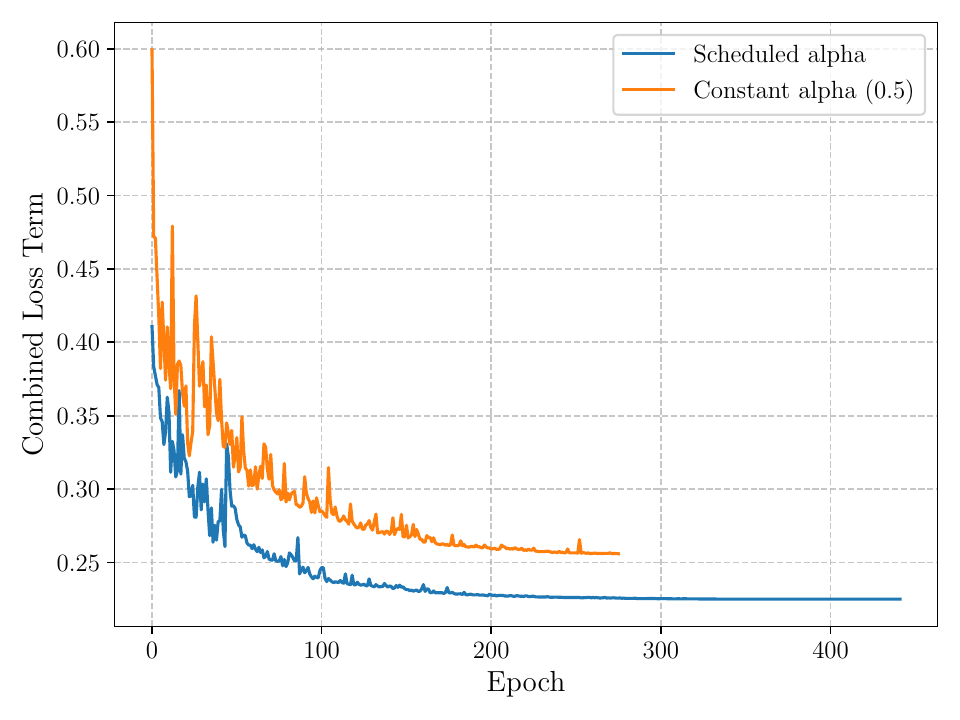}
    \caption{Comparison between total loss term values for two techniques, a constant scheduler with $\alpha = 0.5$ and a warmup scheduler starting at $\alpha_0 = 0.001$ increasing to $\alpha_{\mathrm{max}} = 0.75$ over 200 epochs.}
    \label{fig:performance-alpha}
\end{figure}

Results of experiments performed with gradual increase of $\alpha$ and no warmup demonstrate a lower volatility of training at lower epochs, and an increased expressiveness of the network at later stages of training, visualized in Figure \ref{fig:performance-alpha}. These trends match with the initial observations of the effect of high and low alpha values on training.

\subsection{Runtime}

The addition of the complex automatic differentiation required to calculate and propagate the physics-informed loss term adds significant computational complexity to PINNs when compared to DNNs with an equivalent architecture.

\begin{figure}[H]
    \centering
    \includegraphics[width=\linewidth]{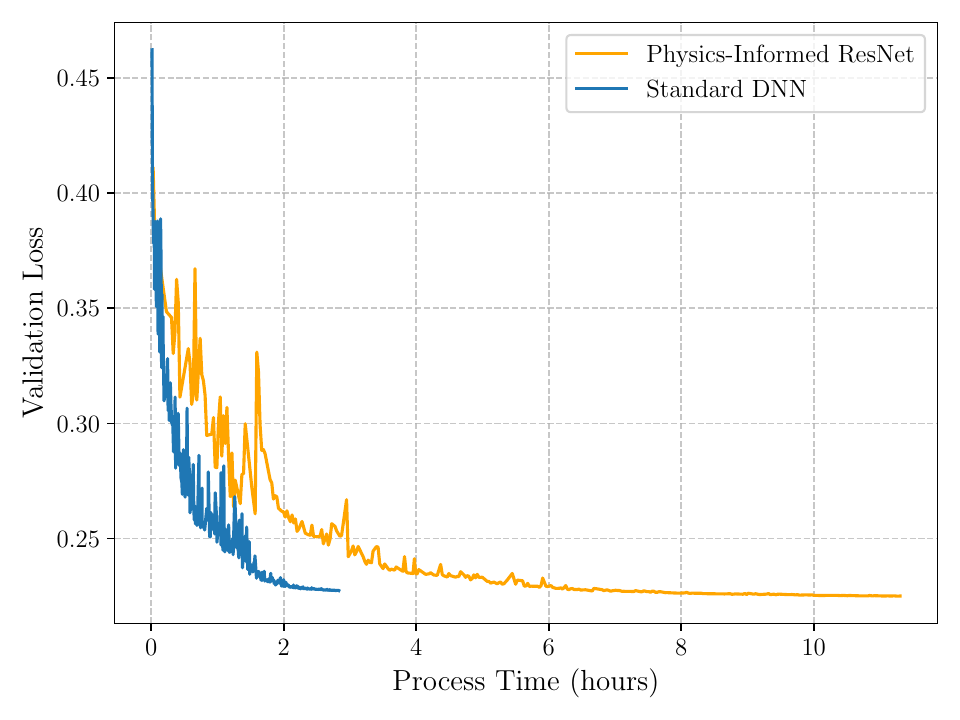}
    \caption{Comparison of training loss as a function of relative wall time between standard DNN network, and PINN architecture.}
    \label{fig:relative wall time}
\end{figure}

Results indicate that PINNs require an average of $\sim2$ additional hours of training when compared to comparable DNNs to reach similar results of $\mathcal{L}_u$.

During training, computations slow down from circa 100 iterations per second to just 30 iteration per second, which indicates a slowdown of the order of $300\%$ at training.  However, it must be noted that as loss terms are not evaluated during inference both DNNs and PINNs are equivalent at prediction time.

    \section{Results}

In the following section we will present results obtained regarding the application of physics-informed neural networks to the three-body problem. Additionally, a comparative analysis between baseline models and physics-informed counterparts will be given.

\subsection{Physics-Informed Error}

The physics-informed error can be seen as a metric of adherence of the model to the physical laws governing a given system. This implies that models with lesser errors are more coherent with reality, and are thus one step further to becoming a reliable solver for the three-body problem.

\begin{figure}[H]
    \centering
    \includegraphics[width=\linewidth]{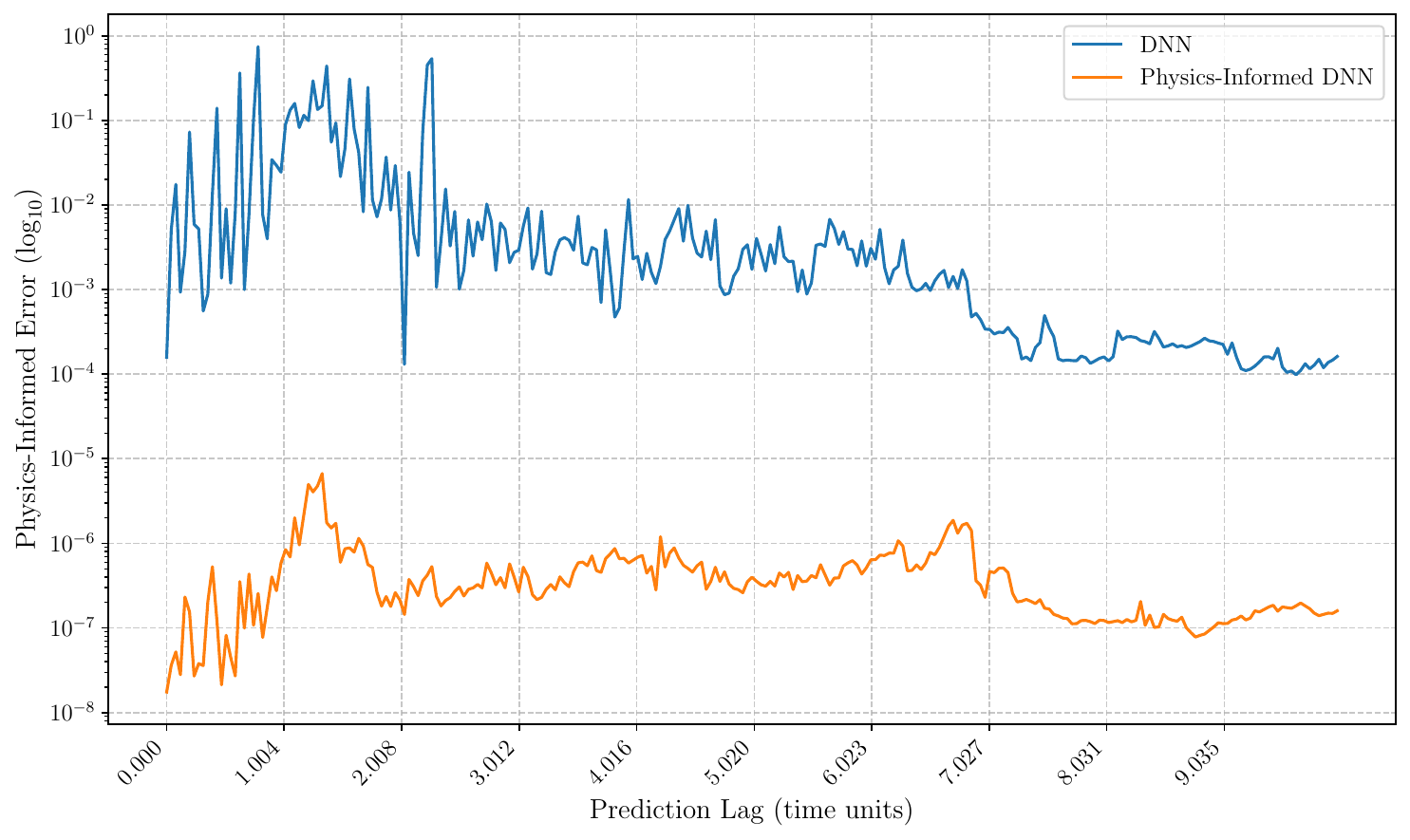}
    \caption{Log-scaled comparison of physics-informed error between the baseline DNN architecture and its physics-informed counterpart.}
    \label{fig:pi-error-dnn}
\end{figure}

Figure \ref{fig:pi-error-dnn} presents an overview of the physics-informed error of the baseline DNN model and its physics-informed counterpart with respect to the prediction lag, \emph{ie.} the timestep predicted for.

Results indicate that the inclusion of the physics-informed (PI) loss during training leads to a significant decrease in the PI error, as the PI-DNN model reaches errors of the order of $1\times10^{-8}$. 

As such, the following figure presents the physics-informed errors achieved by the ResNet architectures, both baseline and PI.

\begin{figure}[H]
    \centering
    \includegraphics[width=\linewidth]{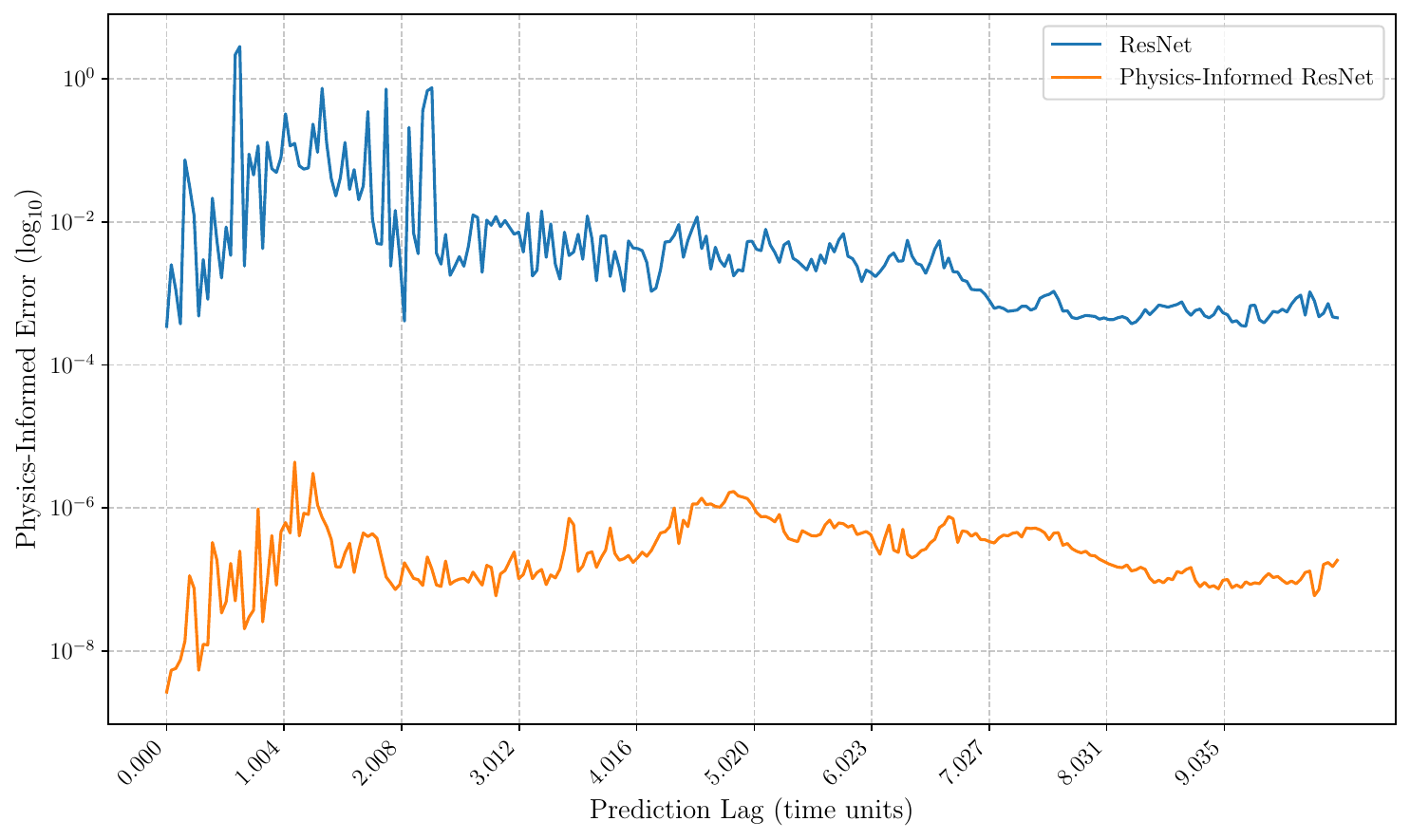}
    \caption{Log-scaled comparison of physics-informed error between the baseline ResNet architecture and its physics-informed counterpart.}
    \label{fig:pi-error-resnet}
\end{figure}

Once again, results demonstrate how the addition of the physics-informed prior leads to a significant decrease in the physics coherence error.

\begin{table*}[t!]
    \centering
    \caption{Error results for non-autoregressive architectures.}
    \label{tab:ablation-non-autoregressive}
    \begin{tabularx}{\textwidth}{
     >{\raggedright\arraybackslash}p{0.17\linewidth}
     !{\vrule width 1.5pt}
     *{4}{>{\centering\arraybackslash}X}
     }
    \toprule
     Model & MAE & RMSE & SMAPE\\
    \midrule
     Baseline DNN & $0.269\,(\pm 0.012)$ & $0.559\,(\pm 0.013)$ & $33.810\,(\pm1.813)\%$\\
     PI DNN & $0.276\,(\pm 0.004)$ & $0.566\,(\pm 0.005)$  & $34.891\,(\pm0.373)\%$\\
     ResNet & $\mathbf{0.243\,(\pm 0.001)}$ & $\mathbf{0.528\,(\pm 0.001)}$ & $\mathbf{30.055\,(\pm 0.041)}\%$\\
     PI ResNet & $0.252\,(\pm 0.004)$ & $0.541\,(\pm 0.006)$ & $31.860\, (\pm0.787)\%$\\
    \bottomrule
    \end{tabularx}
\end{table*}

\begin{table}[H]
    \centering
    \renewcommand{\arraystretch}{1.5}
    \caption{Comparison of physics-informed error values obtained by DNN and ResNet models and their physics-informed counterparts.}
    \label{tab:pi-error}
    \begin{tabularx}{\linewidth}{
     >{\raggedright\arraybackslash}p{0.3\linewidth}
     !{\vrule width 1.5pt}
     *{1}{>{\centering\arraybackslash}X}
     }
    \toprule
     Model & Mean-Square Physics-Informed Error \\
    \midrule
    Baseline DNN & $0.028\,(\pm 0.022)$\\
    PI DNN & $1.825\,(\pm 1.192)\times10^{-6}$\\
    ResNet & $0.030\,(\pm 0.011)$\\
    PI ResNet & $\mathbf{3.291}\,(\pm \mathbf{0.493})\times10^{-7}$\\
    \bottomrule
    \end{tabularx}
\end{table}

Results indicate that despite similar error values according to the data-localization MAE loss (Equation \eqref{eq:pinn-Lu-mae}), physics-informed errors significantly differ between standard models and their physics-informed counterparts, as shown in Table \ref{tab:pi-error}. Notably, the PI ResNet model achieves the lowest physics-informed error among all models tested.

\subsection{Quality of inference}

This subsection will provide a qualitative and quantitative analysis of the quality of inference provided by different models, both physics-informed and not.

\begin{figure}[htbp!]
    \centering
    \includegraphics[width=\linewidth]{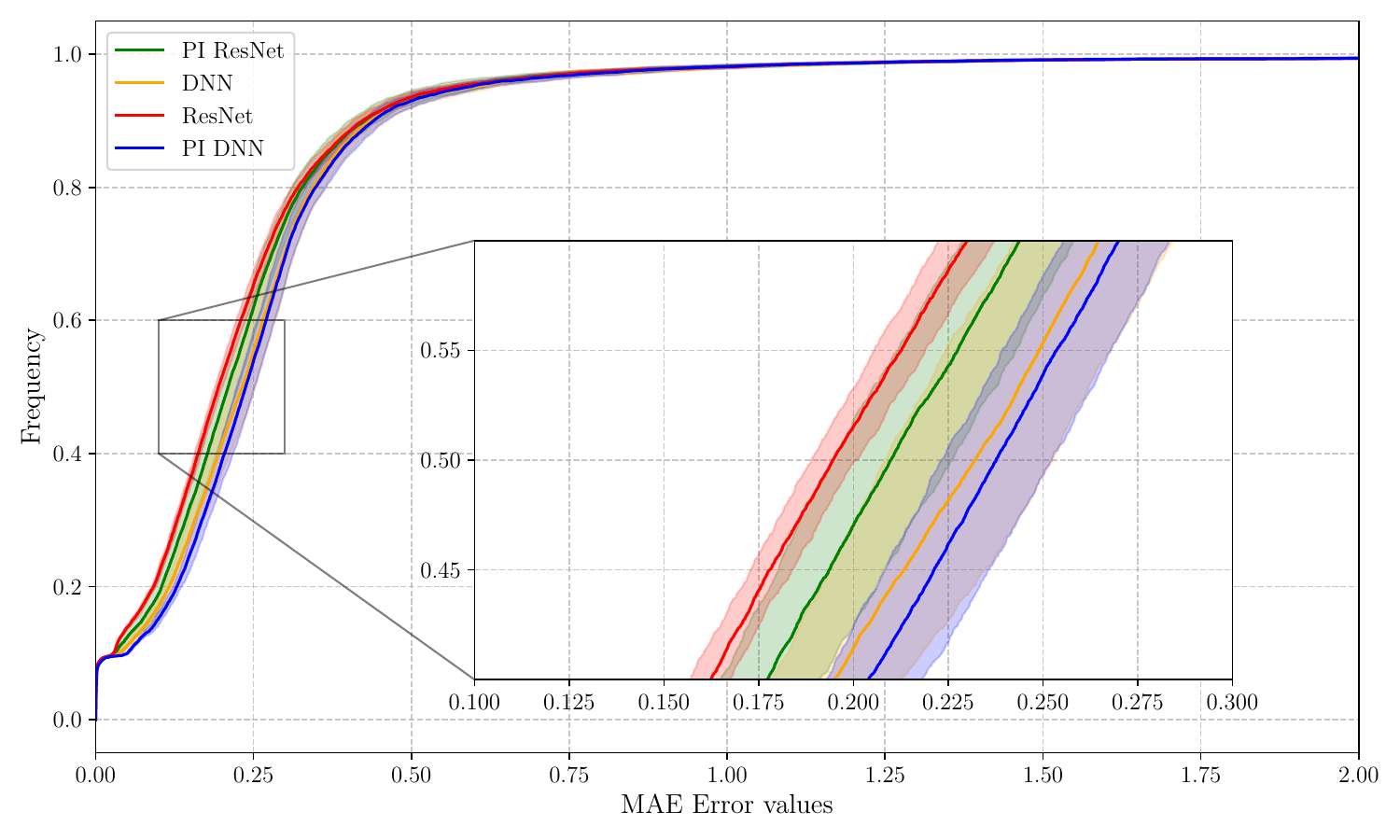}
    \caption{Empirical cumulative distribution function (ECDF) comparison across all models, including uncertainty estimates.}    
    \label{fig:ecdf-model-comparison}
\end{figure}

The empirical cumulative distribution function (ECDF) comparison with respect to the MAE, visualized in Figure \ref{fig:ecdf-model-comparison} displays the difference in performance of all models, with the ResNet models outperforming the DNN, and within each group, the physics-informed model perfoming worst. 
Additionally, the results illustrate a low variance of the error across different samples, with $\approx 99\%$ of samples reporting an MAE value under 2.

Taking into account Table \ref{tab:ablation-non-autoregressive}, the given values prove results discussed until now, demonstrating that across all metrics the standard ResNet model has an increased performance.

\clearpage

\begin{figure*}[!p]  %
    \centering
    \begin{subfigure}[t]{0.58\linewidth}
        \centering
        \includegraphics[width=\textwidth]{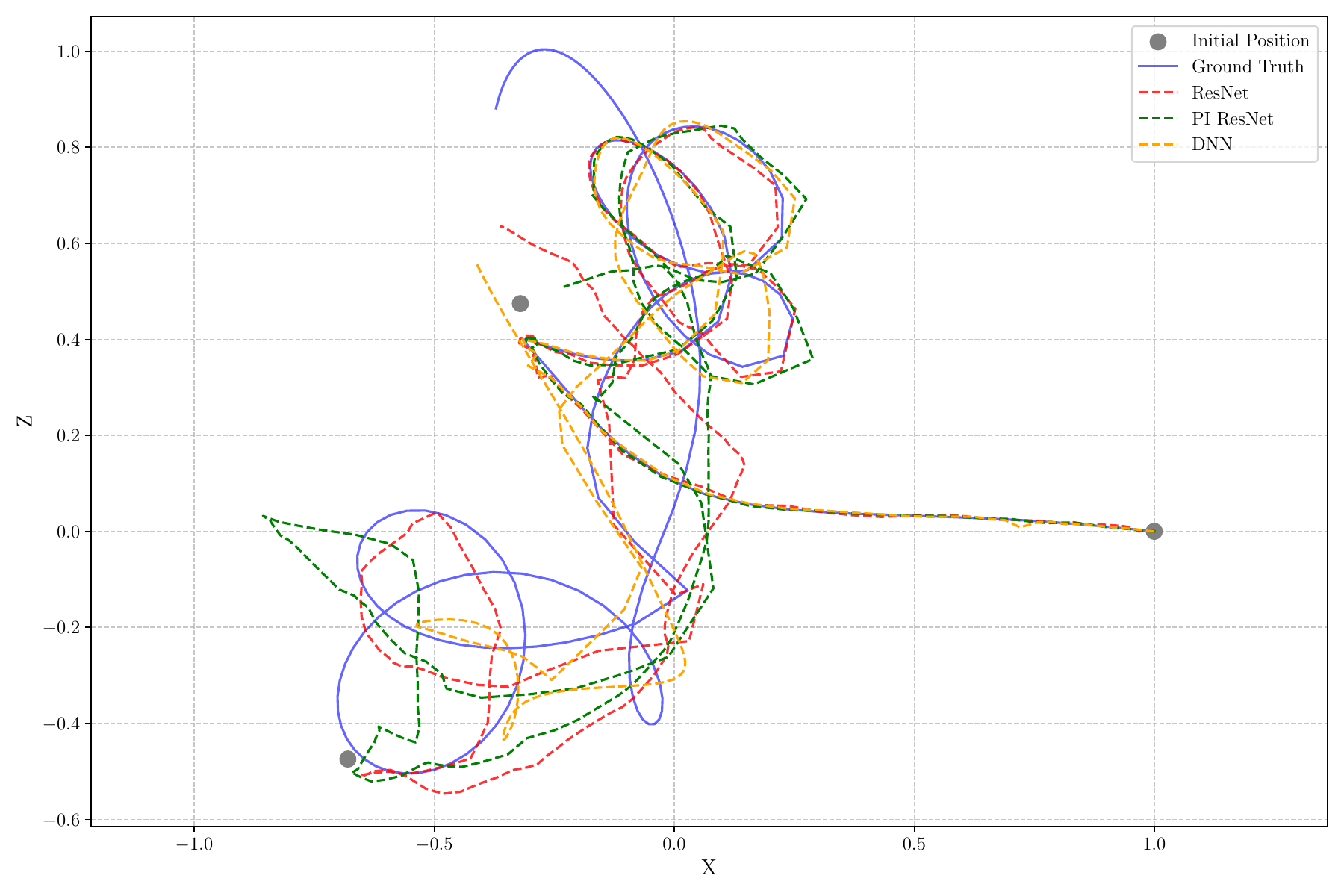}
        \caption{Trajectory of particle 1.}
    \end{subfigure}
    \hfill
    \begin{subfigure}[t]{0.39\linewidth}
        \centering
        \raisebox{0.15\height}{  %
            \includegraphics[width=\textwidth]{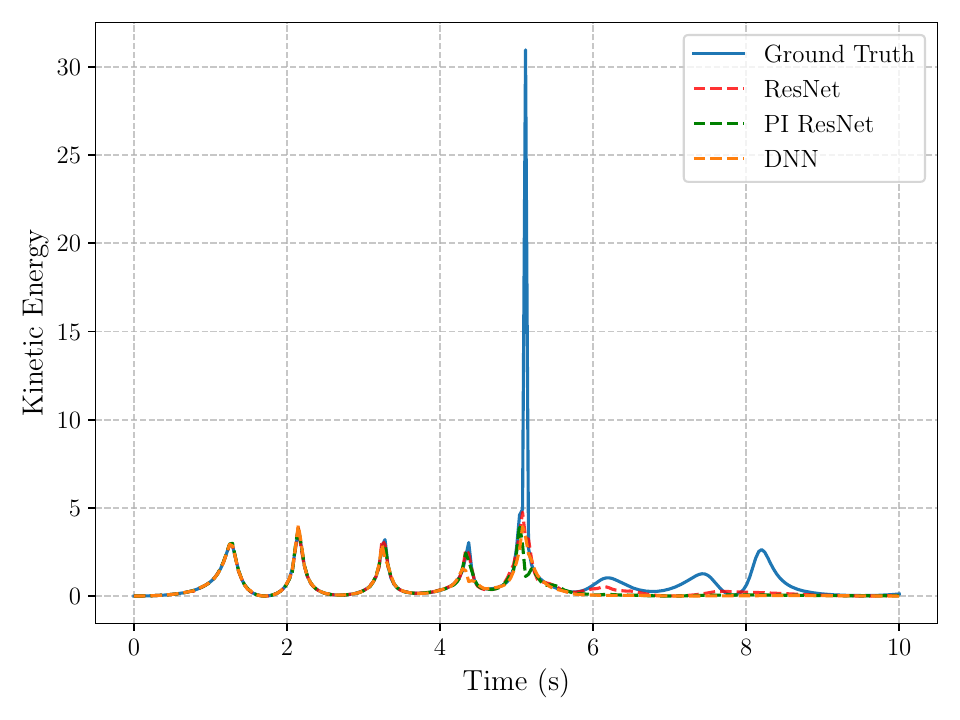}
        }
        \caption{Kinetic energy of particle 1.}
    \end{subfigure}

    \vspace{1em}  %
    \begin{subfigure}[t]{0.58\linewidth}
        \centering
        \includegraphics[width=\textwidth]{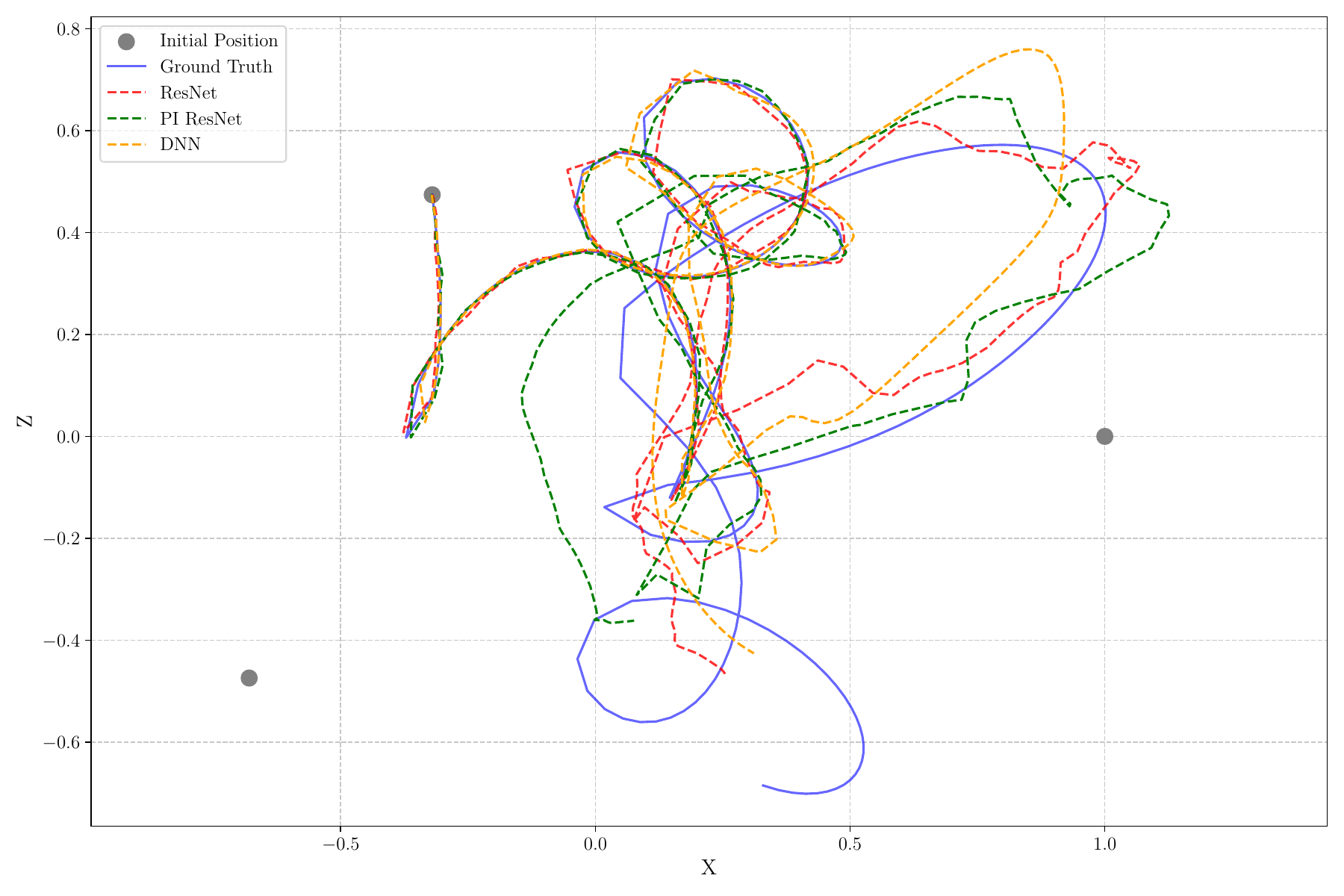}
        \caption{Trajectory of particle 2.}
    \end{subfigure}
    \hfill
    \begin{subfigure}[t]{0.39\linewidth}
        \centering
        \raisebox{0.15\height}{
            \includegraphics[width=\textwidth]{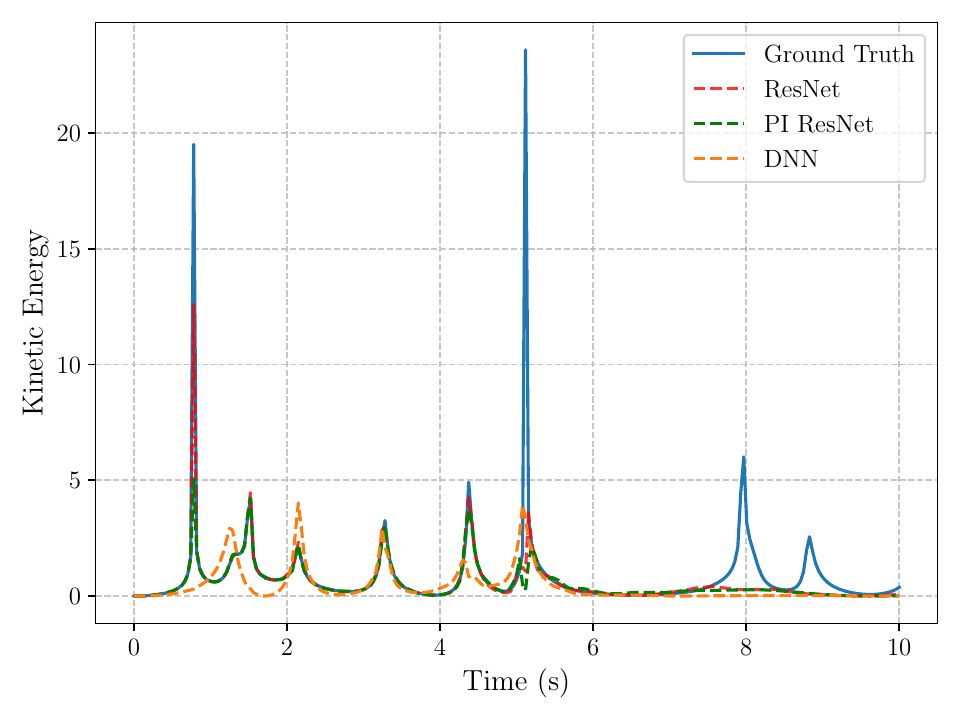}
        }
        \caption{Kinetic energy of particle 2.}
    \end{subfigure}

    \vspace{1em}  %
    \begin{subfigure}[t]{0.58\linewidth}
        \centering
        \includegraphics[width=\textwidth]{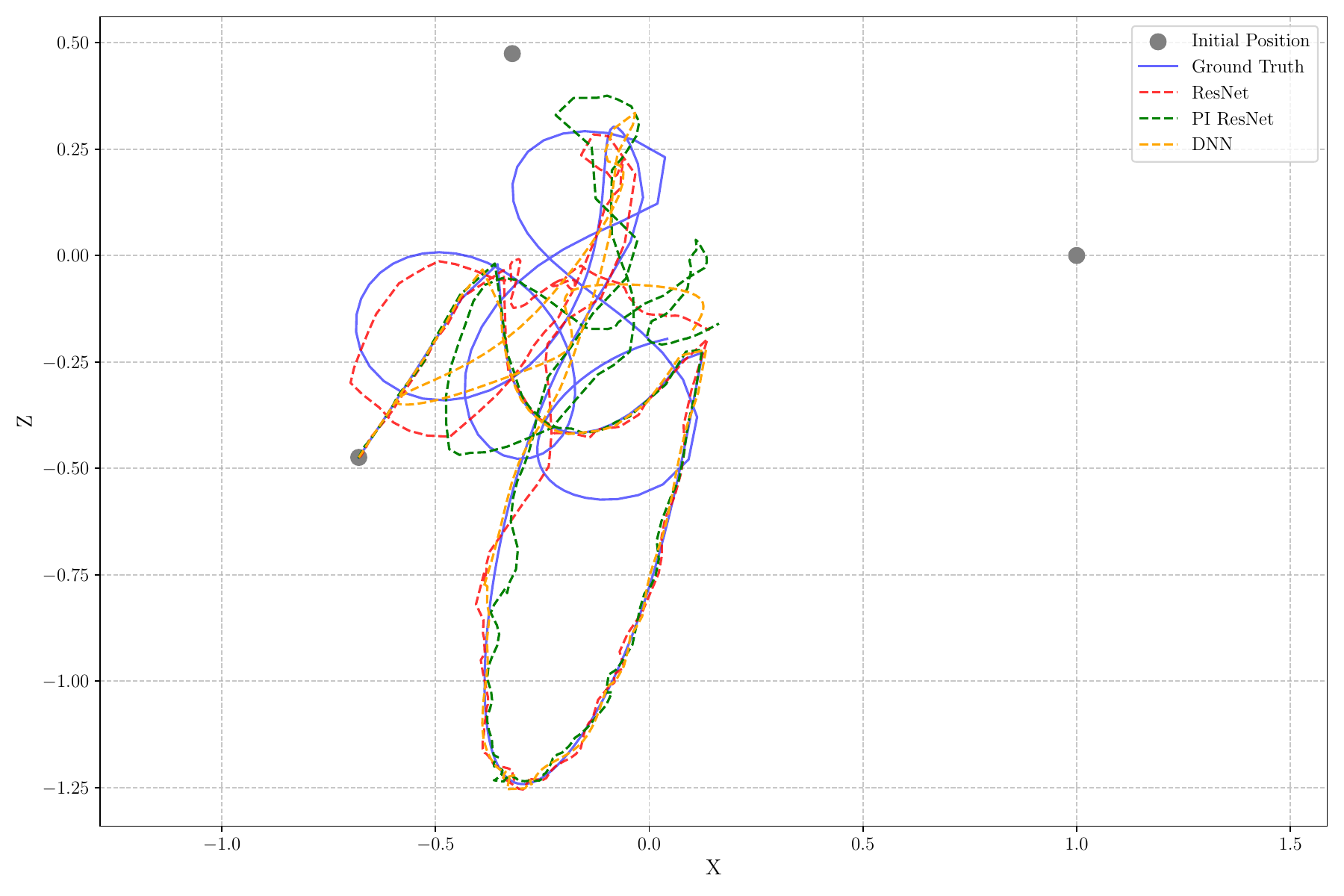}
        \caption{Trajectory of particle 3.}
    \end{subfigure}
    \hfill
    \begin{subfigure}[t]{0.39\linewidth}
        \centering
        \raisebox{0.15\height}{
            \includegraphics[width=\textwidth]{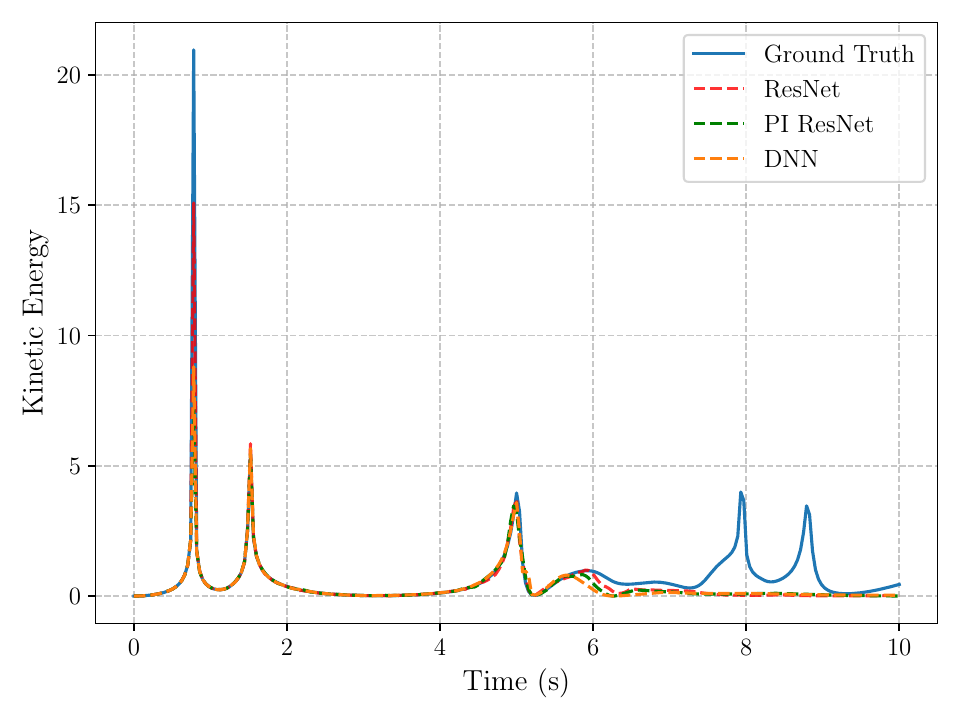}
        }
        \caption{Kinetic energy of particle 3.}
    \end{subfigure}
    
    \caption{Visualization of the quality of predictions made by the best seed of each network architecture for a \textbf{typical system}. For visual clarity, only the ResNet, Physics-Informed (PI) ResNet and DNN model were plotted.}
    \label{fig:trajectories-typical}
\end{figure*}

\clearpage

\begin{figure*}[!p]  %
    \centering
    \begin{subfigure}[t]{0.58\linewidth}
        \centering
        \includegraphics[width=\textwidth]{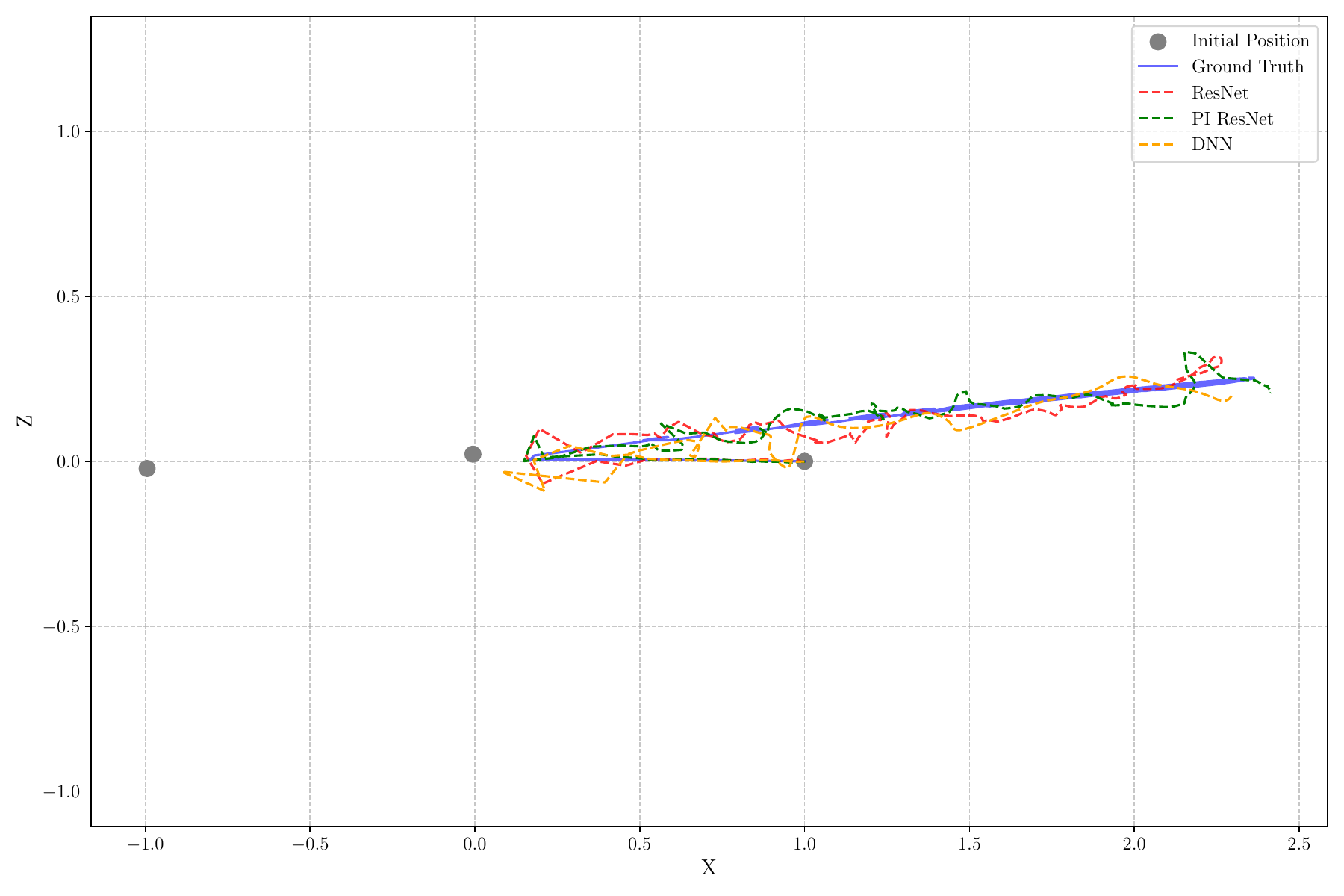}
        \caption{Trajectory of particle 1.}
    \end{subfigure}
    \hfill
    \begin{subfigure}[t]{0.39\linewidth}
        \centering
        \raisebox{0.15\height}{  %
            \includegraphics[width=\textwidth]{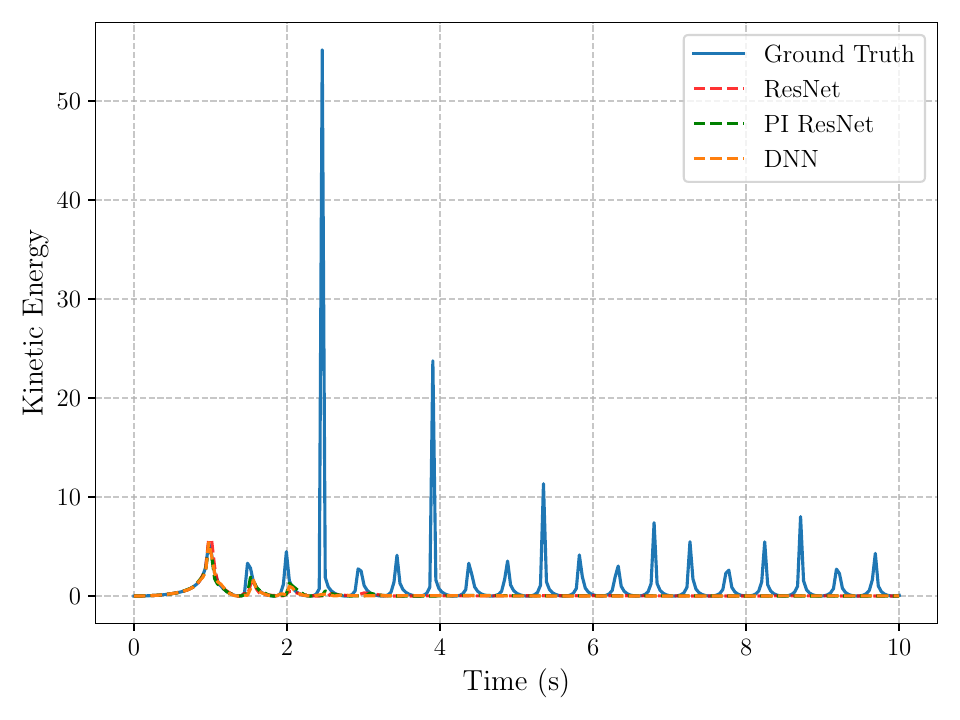}
        }
        \caption{Kinetic energy of particle 1.}
    \end{subfigure}

    \vspace{1em}  %
    \begin{subfigure}[t]{0.58\linewidth}
        \centering
        \includegraphics[width=\textwidth]{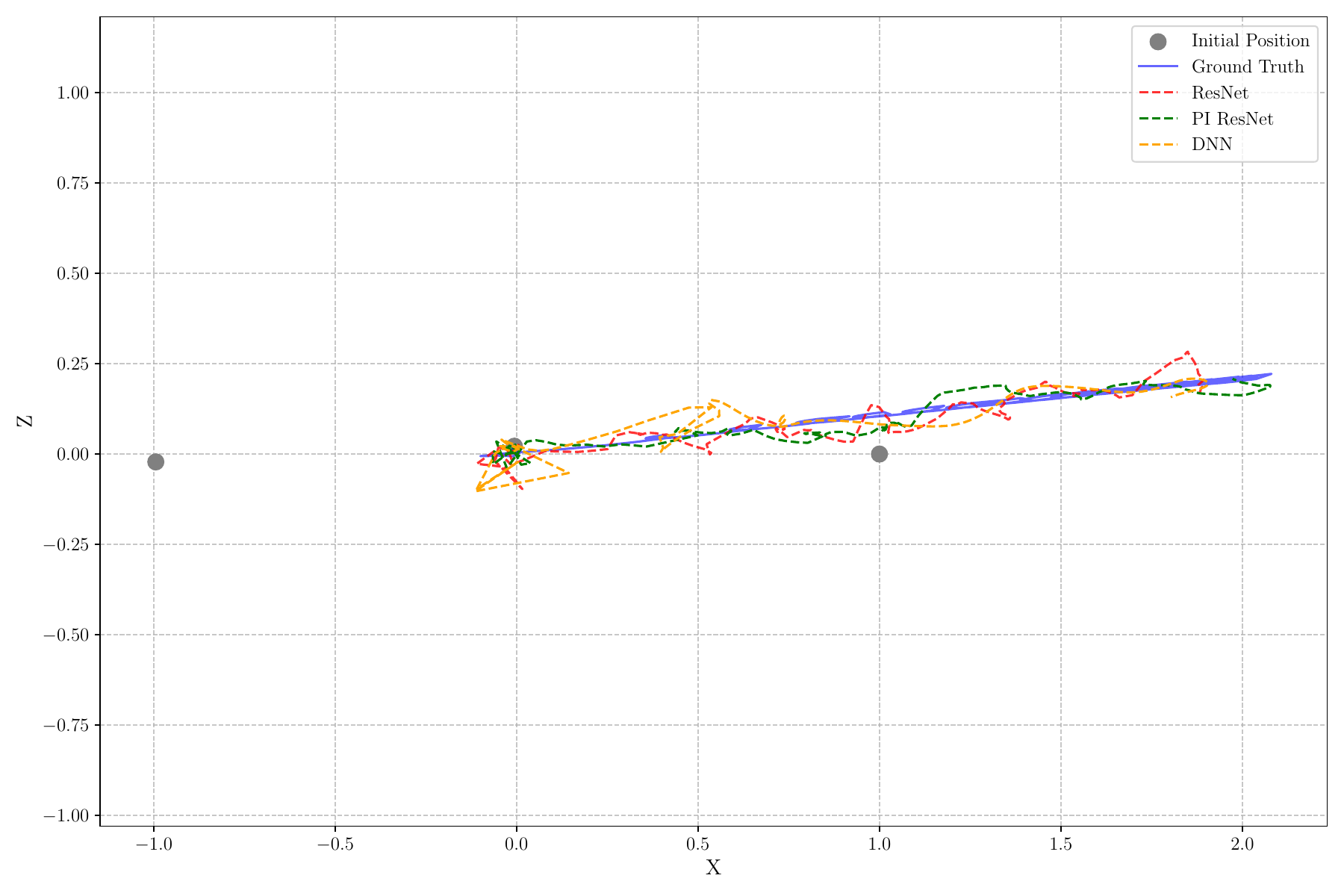}
        \caption{Trajectory of particle 2.}
    \end{subfigure}
    \hfill
    \begin{subfigure}[t]{0.39\linewidth}
        \centering
        \raisebox{0.15\height}{
            \includegraphics[width=\textwidth]{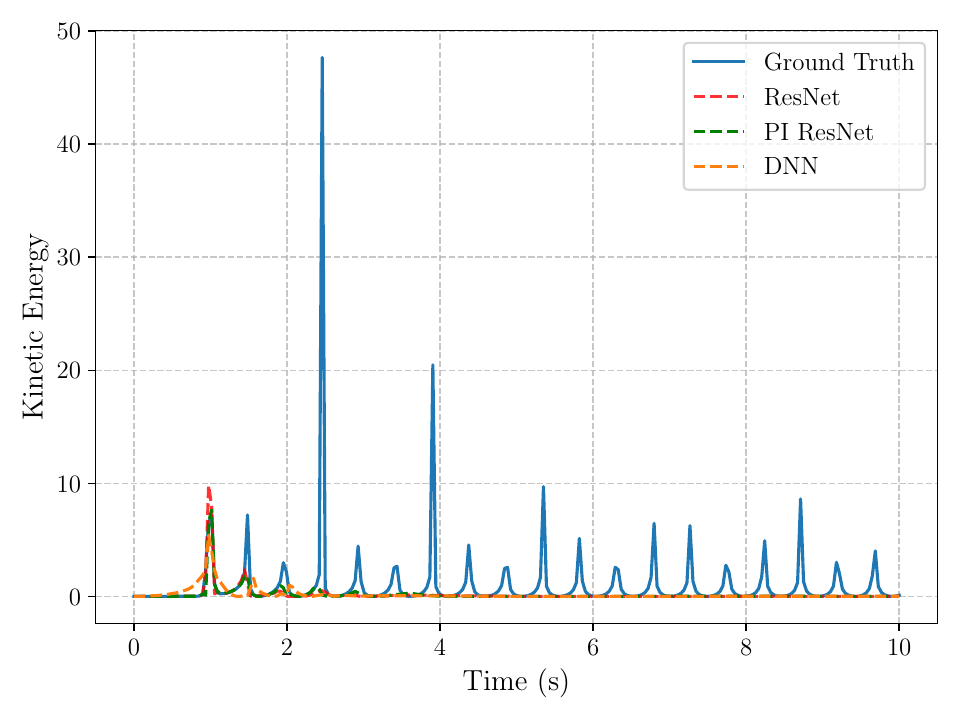}
        }
        \caption{Kinetic energy of particle 2.}
    \end{subfigure}

    \vspace{1em}  %
    \begin{subfigure}[t]{0.58\linewidth}
        \centering
        \includegraphics[width=\textwidth]{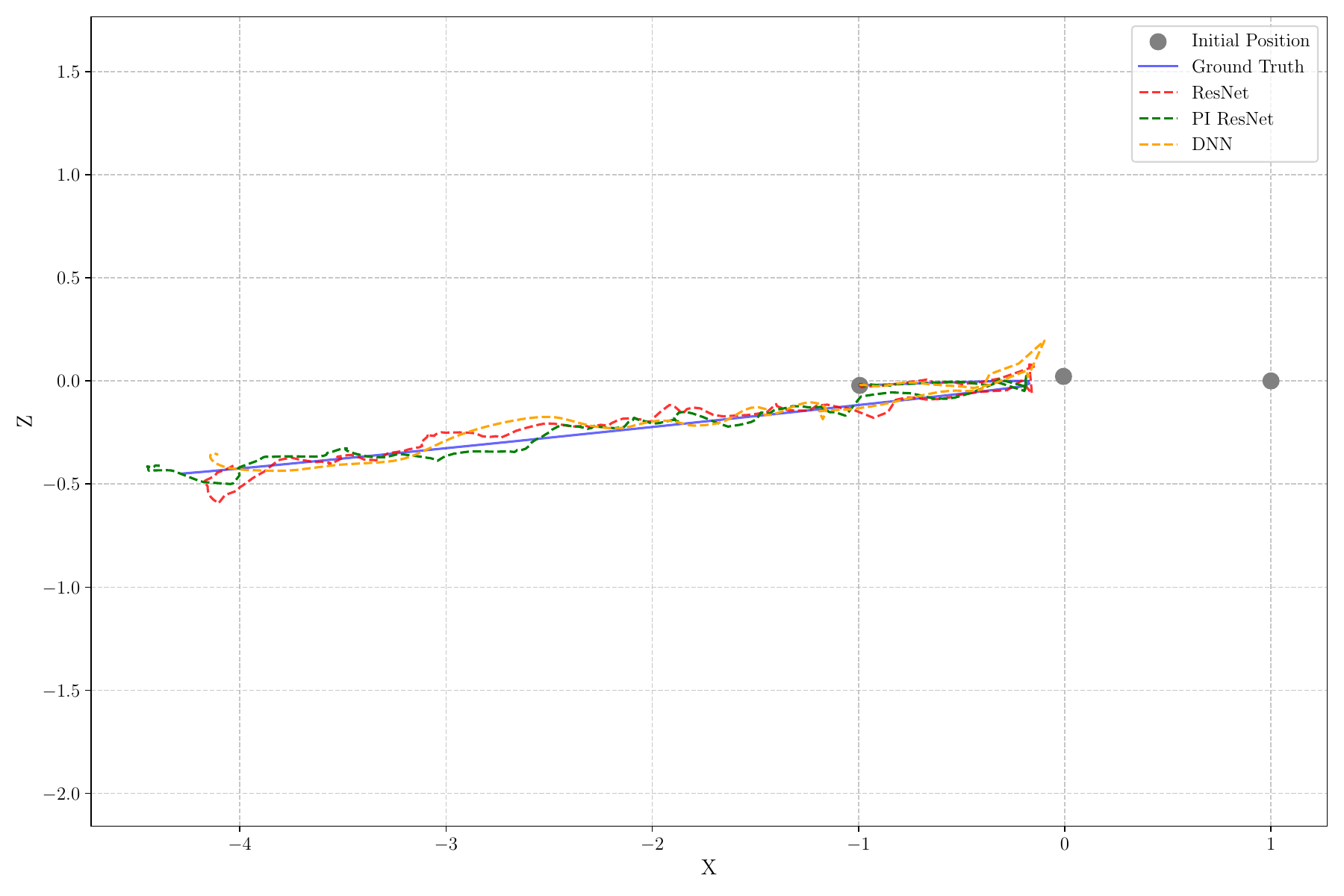}
        \caption{Trajectory of particle 3.}
    \end{subfigure}
    \hfill
    \begin{subfigure}[t]{0.39\linewidth}
        \centering
        \raisebox{0.15\height}{
            \includegraphics[width=\textwidth]{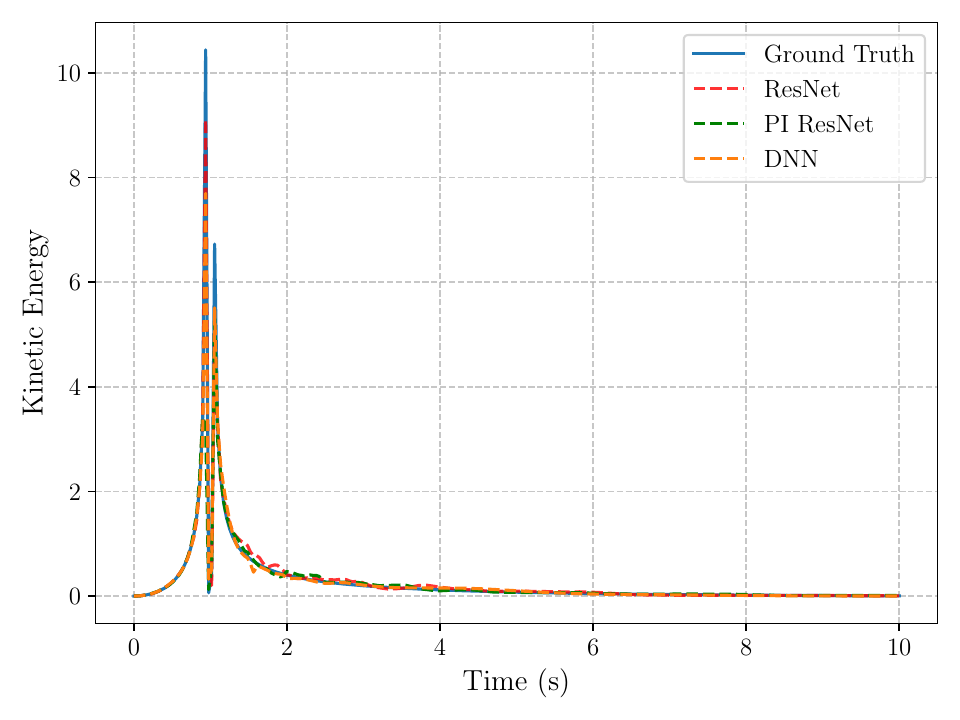}
        }
        \caption{Kinetic energy of particle 3.}
    \end{subfigure}
    
    \caption{Visualization of the quality of predictions made by the best seed of each network architecture for a \textbf{system with near-collinear initial positions}. For visual clarity, only the ResNet, Physics-Informed (PI) ResNet and DNN model were plotted.}
    \label{fig:trajectories-atypical}
\end{figure*}

\clearpage

\subsection{Analysis of predictions}
\label{subsec:prediction-analysis}

This subsection will provide a succint analysis of observations on the predicted orbits generated by the ResNet, physics-informed (PI) ResNet and DNN models. Specifically, the Figures \ref{fig:trajectories-typical} and \ref{fig:trajectories-atypical} will be reviewed.

The system visualized in Figure \ref{fig:trajectories-typical} can be viewed as a representation of a typical system due to its relatively evenly-spaced initial positions and the lack of an ejection of a body from the system within the simulated timeframe. The first observation to be made is the existence of visual disconnects in the expected movement of bodies at specific positions.

\begin{figure}[htbp]
    \centering
    \includegraphics*[width=\linewidth]{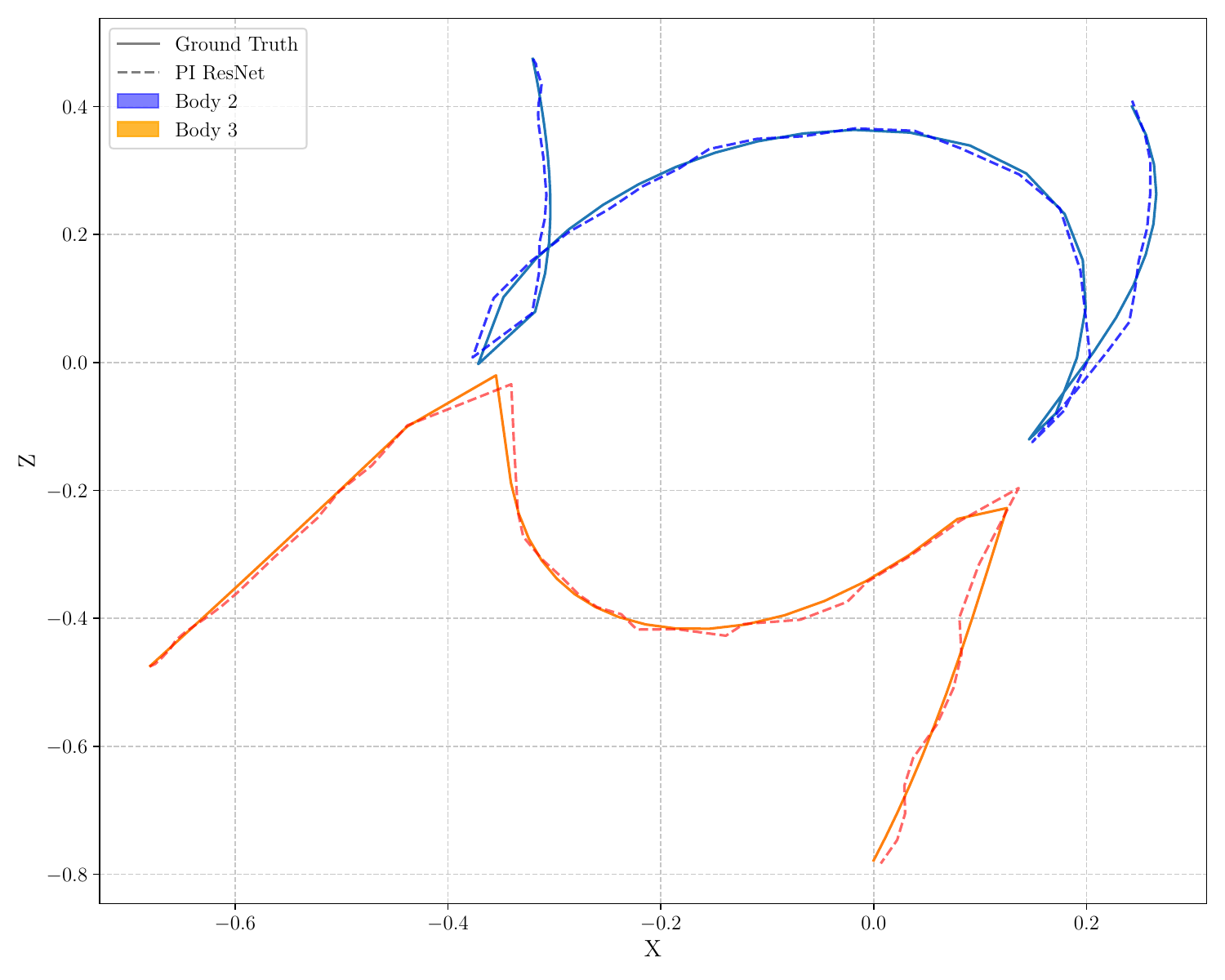}
    \caption{Example of disconnects in object trajectory caused by roughness of simulation timesteps, visualized through zoom-in of Figure \ref{fig:trajectories-typical} at a limited timeframe of $\mathrm{t} = 1.95$ time units.}
    \label{fig:disconnect}
\end{figure}

These disconnects seem to be located near areas of high acceleration, specifically within the frame of a near-encounter by two bodies not moving in parallel. Within Figure \ref{fig:trajectories-typical}, the disconnects seem to be roughly located at times $\mathrm{t_0} = 0.78$ and $\mathrm{t_1} = 1.52$ time units between body 2 and 3. Despite an acceptable fit of the predicted orbits to the ground truth, the kinetic energy indicates issues with the predictions otherwise unseen. 

By analysing the kinetic energies of all models at the relevant timesteps, $t_0$ and $t_1$, we see that physics-informed models tend to under-perform  at these locations, providing significantly more erroneous predictions. Taking these energies into account, predictions seem to remain considerably close to the ground truth until $\mathrm{t} \approx 5$ time units, after which trajectories degenerate and kinetic energies reduce to near-zero values.

On the other hand, by analysing the system displayed in Figure \ref{fig:trajectories-atypical} we may gather how the models react to an atypical system where the initial coordinates are nearly collinear and there is a close encounter by all three bodies simultaneously. Once again there are disconnects where the behaviour of the system is hidden due to the granularity of the simulation timestep. It is at these disconnects, $\mathrm{t}_0 \approx 1$ and $\mathrm{t}_1 \approx 2$ time units where all models cease to predict kinetic energies accurately and eventually degenerate. However, contrary to the typical system, trajectory prediction does not degrade and remains accurate.

These results seem to indicate that high-acceleration close encounter regions are the cause of degenerating predictions. Yet, these results are biased due to the granularity of the simulation, as the acceleration reported by these is more sudden than would be physically realistic.

    \section{Conclusions}

This work has demonstrated how Physics-Informed Neural Networks can be used to enhance the coherence of machine learning predictions to physical laws governing the studied system, even if chaotic in nature. Additionally, it was proved how the ResNet architecture can provide improved prediction capabilites comparatively to the existing DNN baselines. The ability to more accurately provide solutions to this notoriously complex problem at reduced computational costs represents an advancement within the applied study of machine learning to celestial mechanics and dynamical systems.

The success of \glspl{PINN} in offering time-efficient solutions to the three-body problem opens up exciting possibilities to tackle problems within dynamical systems in physics and engineering. 

In conclusion, not only does our work contribute to the growing body of research that supports the application of machine learning to the physical sciences, but also offers insights into the relation of physics-informed machine learning to classical machine learning within complex dynamical problems.

    \section{Future Work}

While this study has demonstrated the capability of physics-informed neural networks to operate under small-data regimes and perform improved learning in the complex dynamical system that defines the three-body problem, several limitations remain that offer possibilities for future work. These limitations include:
\begin{enumerate}
    \item Loss Landscape Constraining: A potential limitation on the learning abilities of \glspl{PINN} is the possibility that the physics-informed loss term induces constraints harsh enough to effectively impede learning. Future work would be required to analyze the need for, and propose possible mitigation strategies.
    \item Long-term Predictions: Within this work we have proved the success of \glspl{PINN} within a time interval of 10 seconds. However, due to the chaotic nature of the \gls{TBP} no assertions can be made regarding longer timescales, and studies must be conducted on the limits of prediction.
    \item Step Precision: All networks in this work have been trained on timesteps of $0.0390625$ seconds, and no assertions can be made regarding the ability of the network to predict for non-multiples of the aforementioned step size. Additionally, in Subsection \ref{subsec:prediction-analysis}, we have viewed how disconnects in simulated data caused by coarse timesteps may lead to a degrading of predictions. Future work would deepen the study of the this relation, and seek methods to minimize these issues. 
\end{enumerate}

Future research could focus not only on adressing these mentioned standing issues, but also in the application of newer and promising architectures and methodologies. Several machine learning frameworks were identified as offering potential developments:
\begin{enumerate}
    \item Hamiltonian Neural Networks \cite{Greydanus2019}
    \item Lagrangian Neural Networks \cite{Cranmer2020}
    \item Neural Operators:
    \begin{enumerate}
        \item Fourier Neural Operators \cite{Li2021}
        \item Convolutional Neural Operators \cite{Raonic2024}
    \end{enumerate}
\end{enumerate} 

These advanced architectures could potentially increase the performance of neural networks within the context of the three-body problem, making use of alternate formal definitions of the problem such as the Hamiltonian and Lagrangian formalisms. By exploring their integration with \glspl{PINN} or comparing their performance on existing periodic solutions, valuable insights could be gained and advancements in the field made.

    \section*{Acknowledgements}

This research was carried out under Project “Artificial Intelligence Fights Space Debris” Nº C626449889-0046305 co-funded by Recovery and Resilience Plan and NextGeneration EU Funds, \url{www.recuperarportugal.gov.pt}.

    \bibliographystyle{unsrtnat}
    \bibliography{references}

\begin{thebibliography}{37}
\providecommand{\natexlab}[1]{#1}
\providecommand{\url}[1]{\texttt{#1}}
\expandafter\ifx\csname urlstyle\endcsname\relax
  \providecommand{\doi}[1]{doi: #1}\else
  \providecommand{\doi}{doi: \begingroup \urlstyle{rm}\Url}\fi

\bibitem[{Newton}(1687)]{Newton1687}
Isaac {Newton}.
\newblock \emph{{Philosophiae Naturalis Principia Mathematica.}}
\newblock 1687.
\newblock \doi{10.3931/e-rara-440}.

\bibitem[Euler(1767)]{Euler1767}
Leonhard Euler.
\newblock De motu rectilineo trium corporum se mutuo attrahentium.
\newblock \emph{Novi commentarii academiae scientiarum Petropolitanae}, pages 144--151, 1767.

\bibitem[Lagrange(1772)]{Lagrange1772}
Joseph-Louis Lagrange.
\newblock Essai sur le probleme des trois corps.
\newblock \emph{Prix de l’acad{\'e}mie royale des Sciences de paris}, 9:\penalty0 292, 1772.

\bibitem[Poincar{\'e}(1893)]{Poincare1893}
Henri Poincar{\'e}.
\newblock \emph{Les m{\'e}thodes nouvelles de la m{\'e}canique c{\'e}leste}, volume~2.
\newblock Gauthier-Villars et fils, imprimeurs-libraires, 1893.

\bibitem[Sundman(1909)]{Sundman1909}
K~Sundman.
\newblock Le probl{\`e}me des trois corps.
\newblock \emph{Acta Soc. Sci. Fenn}, 35:\penalty0 1909, 1909.

\bibitem[Broucke and Lass(1973)]{Broucke1973}
R~Broucke and H~Lass.
\newblock A note on relative motion in the general three-body problem.
\newblock \emph{Celestial mechanics}, 8\penalty0 (1):\penalty0 5--10, 1973.

\bibitem[Šuvakov and Dmitrašinović(2013-03)]{Suvakov2013}
Milovan Šuvakov and V.~Dmitrašinović.
\newblock Three classes of newtonian three-body planar periodic orbits.
\newblock \emph{Physical Review Letters}, 110\penalty0 (11), 2013-03.
\newblock ISSN 1079-7114.
\newblock \doi{10.1103/physrevlett.110.114301}.

\bibitem[Moore(1993)]{Moore1993}
Cristopher Moore.
\newblock Braids in classical dynamics.
\newblock \emph{Physical Review Letters}, 70\penalty0 (24):\penalty0 3675, 1993.

\bibitem[Li and Liao(2017)]{Li2017}
XiaoMing Li and ShiJun Liao.
\newblock More than six hundred new families of newtonian periodic planar collisionless three-body orbits.
\newblock \emph{Science China Physics, Mechanics \& Astronomy}, 60:\penalty0 1--7, 2017.

\bibitem[Liao and Li(2019-07)]{Liao2019}
Shijun Liao and Xiaoming Li.
\newblock {On the periodic solutions of the three-body problem}.
\newblock \emph{National Science Review}, 6\penalty0 (6):\penalty0 1070--1071, 2019-07.
\newblock ISSN 2095-5138.
\newblock \doi{10.1093/nsr/nwz102}.

\bibitem[Levien and Tan(1993-11)]{Levien1993}
R.~B. Levien and S.~M. Tan.
\newblock {Double pendulum: An experiment in chaos}.
\newblock \emph{American Journal of Physics}, 61\penalty0 (11):\penalty0 1038--1044, 1993-11.
\newblock ISSN 0002-9505.
\newblock \doi{10.1119/1.17335}.

\bibitem[Musielak and Quarles(2014-06)]{Musielak2014}
Z~E Musielak and B~Quarles.
\newblock The three-body problem.
\newblock \emph{Reports on Progress in Physics}, 77\penalty0 (6):\penalty0 065901, 2014-06.
\newblock ISSN 1361-6633.
\newblock \doi{10.1088/0034-4885/77/6/065901}.

\bibitem[Carleo et~al.(2019)Carleo, Cirac, Cranmer, Daudet, Schuld, Tishby, Vogt-Maranto, and Zdeborov{\'a}]{Carleo2019}
Giuseppe Carleo, Ignacio Cirac, Kyle Cranmer, Laurent Daudet, Maria Schuld, Naftali Tishby, Leslie Vogt-Maranto, and Lenka Zdeborov{\'a}.
\newblock Machine learning and the physical sciences.
\newblock \emph{Reviews of Modern Physics}, 91\penalty0 (4):\penalty0 045002, 2019.

\bibitem[Bishop and Bishop(2023)]{Bishop2023}
Christopher~M Bishop and Hugh Bishop.
\newblock \emph{Deep learning: Foundations and concepts}.
\newblock Springer Nature, 2023.

\bibitem[Raissi and Karniadakis(2018)]{Raissi2018}
Maziar Raissi and George~Em Karniadakis.
\newblock Hidden physics models: Machine learning of nonlinear partial differential equations.
\newblock \emph{Journal of Computational Physics}, 357:\penalty0 125--141, 2018.
\newblock ISSN 0021-9991.
\newblock \doi{https://doi.org/10.1016/j.jcp.2017.11.039}.

\bibitem[Pathak et~al.(2018-01)Pathak, Hunt, Girvan, Lu, and Ott]{Pathak2018}
Jaideep Pathak, Brian Hunt, Michelle Girvan, Zhixin Lu, and Edward Ott.
\newblock Model-free prediction of large spatiotemporally chaotic systems from data: A reservoir computing approach.
\newblock \emph{Phys. Rev. Lett.}, 120:\penalty0 024102, 2018-01.
\newblock \doi{10.1103/PhysRevLett.120.024102}.

\bibitem[Cranmer et~al.(2020)Cranmer, Greydanus, Hoyer, Battaglia, Spergel, and Ho]{Cranmer2020}
Miles Cranmer, Sam Greydanus, Stephan Hoyer, Peter Battaglia, David Spergel, and Shirley Ho.
\newblock Lagrangian neural networks, 2020.

\bibitem[Greydanus et~al.(2019)Greydanus, Dzamba, and Yosinski]{Greydanus2019}
Samuel Greydanus, Misko Dzamba, and Jason Yosinski.
\newblock Hamiltonian neural networks.
\newblock In H.~Wallach, H.~Larochelle, A.~Beygelzimer, F.~d\textquotesingle Alch\'{e}-Buc, E.~Fox, and R.~Garnett, editors, \emph{Advances in Neural Information Processing Systems}, volume~32. Curran Associates, Inc., 2019.

\bibitem[Breen et~al.(2020-04)Breen, Foley, Boekholt, and Zwart]{Breen2020}
Philip~G Breen, Christopher~N Foley, Tjarda Boekholt, and Simon~Portegies Zwart.
\newblock {Newton versus the machine: solving the chaotic three-body problem using deep neural networks}.
\newblock \emph{Monthly Notices of the Royal Astronomical Society}, 494\penalty0 (2):\penalty0 2465--2470, 2020-04.
\newblock ISSN 0035-8711.
\newblock \doi{10.1093/mnras/staa713}.

\bibitem[Fukushima(1975)]{Fukushima1975}
Kunihiko Fukushima.
\newblock Cognitron: A self-organizing multilayered neural network.
\newblock \emph{Biological Cybernetics}, 20\penalty0 (3-4):\penalty0 121--136, 1975.
\newblock \doi{10.1007/bf00342633}.

\bibitem[Lagaris et~al.(1998)Lagaris, Likas, and Fotiadis]{Lagaris1998}
I.E. Lagaris, A.~Likas, and D.I. Fotiadis.
\newblock Artificial neural networks for solving ordinary and partial differential equations.
\newblock \emph{IEEE Transactions on Neural Networks}, 9\penalty0 (5):\penalty0 987--1000, 1998.
\newblock \doi{10.1109/72.712178}.

\bibitem[Raissi et~al.(2019)Raissi, Perdikaris, and Karniadakis]{Raissi2019}
M.~Raissi, P.~Perdikaris, and G.E. Karniadakis.
\newblock Physics-informed neural networks: A deep learning framework for solving forward and inverse problems involving nonlinear partial differential equations.
\newblock \emph{Journal of Computational Physics}, 378:\penalty0 686--707, 2019.
\newblock ISSN 0021-9991.
\newblock \doi{https://doi.org/10.1016/j.jcp.2018.10.045}.

\bibitem[Goodman et~al.(1993)Goodman, Heggie, and Hut]{Goodman1993}
Jeremy Goodman, Douglas~C Heggie, and Piet Hut.
\newblock On the exponential instability of n-body systems.
\newblock \emph{Astrophysical Journal v. 415, p. 715}, 415:\penalty0 715, 1993.

\bibitem[Boekholt et~al.(2014)Boekholt, Boekholt, Zwart, and Zwart]{Boekholt2014}
Tjarda Boekholt, T.~C.~N. Boekholt, Simon~Portegies Zwart, and Simon~Portegies Zwart.
\newblock On the reliability of n-body simulations.
\newblock \emph{arXiv: Instrumentation and Methods for Astrophysics}, 2014.
\newblock \doi{10.1186/s40668-014-0005-3}.

\bibitem[Wisdom and Holman(1992)]{Wisdom1992}
Jack Wisdom and Matthew Holman.
\newblock Symplectic maps for the n-body problem-stability analysis.
\newblock \emph{Astronomical Journal (ISSN 0004-6256), vol. 104, no. 5, p. 2022-2029.}, 104:\penalty0 2022--2029, 1992.

\bibitem[Press(1992)]{Press1992}
William~H Press.
\newblock \emph{Numerical recipes in FORTRAN: the art of scientific computing}.
\newblock Cambridge University Press, 1992.

\bibitem[Runge(1895)]{Runge1895}
C.~Runge.
\newblock Ueber die numerische auflösung von differentialgleichungen.
\newblock \emph{Mathematische Annalen}, 46\penalty0 (2):\penalty0 167--178, 1895.
\newblock ISSN 1432-1807.
\newblock \doi{10.1007/BF01446807}.

\bibitem[Bulirsch and Stoer(1966)]{Bulirsch1966}
Roland Bulirsch and Josef Stoer.
\newblock Numerical treatment of ordinary differential equations by extrapolation methods.
\newblock \emph{Numerische Mathematik}, 8\penalty0 (1):\penalty0 1--13, 1966.
\newblock ISSN 0945-3245.
\newblock \doi{10.1007/BF02165234}.

\bibitem[Kingma and Ba(2017)]{Kingma2017}
Diederik~P. Kingma and Jimmy Ba.
\newblock Adam: A method for stochastic optimization, 2017.

\bibitem[Kumar et~al.(2021)Kumar, Das, and Gupta]{Kumar2021}
Pratyush Kumar, Aishwarya Das, and Debayan Gupta.
\newblock Differential euler: Designing a neural network approximator to solve the chaotic three body problem, 2021.

\bibitem[Mi(2021)]{Mi2021}
Lan Mi.
\newblock Solving the three-body problem using numerical simulations and neural networks.
\newblock In \emph{2021 2nd International Seminar on Artificial Intelligence, Networking and Information Technology (AINIT)}, pages 338--341, 2021.
\newblock \doi{10.1109/AINIT54228.2021.00073}.

\bibitem[Choudhary et~al.(2020-06)Choudhary, Lindner, Holliday, Miller, Sinha, and Ditto]{Choudhary2020}
Anshul Choudhary, John~F. Lindner, Elliott~G. Holliday, Scott~T. Miller, Sudeshna Sinha, and William~L. Ditto.
\newblock Physics-enhanced neural networks learn order and chaos.
\newblock \emph{Phys. Rev. E}, 101:\penalty0 062207, 2020-06.
\newblock \doi{10.1103/PhysRevE.101.062207}.

\bibitem[Sharma et~al.(2023-06)Sharma, Evans, Tindall, and Nithiarasu]{Sharma2023}
Prakhar Sharma, Llion Evans, Michelle Tindall, and Perumal Nithiarasu.
\newblock Stiff-pdes and physics-informed neural networks.
\newblock \emph{Archives of Computational Methods in Engineering}, 30\penalty0 (5):\penalty0 2929--2958, 2023-06.
\newblock ISSN 1886-1784.
\newblock \doi{10.1007/s11831-023-09890-4}.

\bibitem[He et~al.(2016-06)He, Zhang, Ren, and Sun]{He2016}
Kaiming He, Xiangyu Zhang, Shaoqing Ren, and Jian Sun.
\newblock Deep residual learning for image recognition.
\newblock In \emph{Proceedings of the IEEE Conference on Computer Vision and Pattern Recognition (CVPR)}, 2016-06.

\bibitem[Prechelt(1998)]{Prechelt1998}
Lutz Prechelt.
\newblock \emph{Early Stopping - But When?}, pages 55--69.
\newblock Springer Berlin Heidelberg, 1998.
\newblock ISBN 978-3-540-49430-0.
\newblock \doi{10.1007/3-540-49430-8_3}.

\bibitem[Li et~al.(2021)Li, Kovachki, Azizzadenesheli, Liu, Bhattacharya, Stuart, and Anandkumar]{Li2021}
Zongyi Li, Nikola Kovachki, Kamyar Azizzadenesheli, Burigede Liu, Kaushik Bhattacharya, Andrew Stuart, and Anima Anandkumar.
\newblock Fourier neural operator for parametric partial differential equations, 2021.

\bibitem[Raonic et~al.(2024)Raonic, Molinaro, De~Ryck, Rohner, Bartolucci, Alaifari, Mishra, and de~B{\'e}zenac]{Raonic2024}
Bogdan Raonic, Roberto Molinaro, Tim De~Ryck, Tobias Rohner, Francesca Bartolucci, Rima Alaifari, Siddhartha Mishra, and Emmanuel de~B{\'e}zenac.
\newblock Convolutional neural operators for robust and accurate learning of pdes.
\newblock \emph{Advances in Neural Information Processing Systems}, 36, 2024.

\end{thebibliography}

\end{document}